\documentclass[a4paper,fleqn]{cas-dc}
\usepackage[numbers]{natbib}
\usepackage{caption}
\usepackage{floatrow} 
\usepackage{graphicx}
\usepackage{multirow}
\usepackage{makecell}
\usepackage{subfigure}
\usepackage{algorithm}
\usepackage[noend]{algpseudocode}
\usepackage{algorithmicx}
\usepackage{float} 
\def\tsc#1{\csdef{#1}{\textsc{\lowercase{#1}}\xspace}}
\tsc{WGM}
\tsc{QE}

\begin{document}
\let\WriteBookmarks\relax
\def\floatpagepagefraction{1}
\def\textpagefraction{.001}
\let\printorcid\relax 

\shorttitle{}    


\title [mode = title]{Graph Unlearning: Efficient Node Removal in Graph Neural Networks}   

\author[1,2]{Faqian Guan}
\author[1]{Tianqing Zhu\texorpdfstring{\corref{cor1}}{ (Corresponding author)}}\ead{tqzhu@cityu.edu.mo}
\author[2]{Zhoutian Wang}
\author[2]{Wei Ren}
\author[1]{Wanlei Zhou}

\affiliation[1]{organization={Faculty of Data Science, City University of Macau},
            postcode={999078}, 
            state={Macao},
            country={China}}
            
\affiliation[2]{organization={School of Computer Science},
            addressline={China University of Geosciences (Wuhan)}, 
            city={Wuhan},
            postcode={430074}, 
            state={Hubei},
            country={China}}


\begin{abstract}
With increasing concerns about privacy attacks and potential sensitive information leakage, researchers have actively explored methods to efficiently remove sensitive training data and reduce privacy risks in graph neural network (GNN) models. Node unlearning has emerged as a promising technique for protecting the privacy of sensitive nodes by efficiently removing specific training node information from GNN models. However, existing node unlearning methods either impose restrictions on the GNN structure or do not effectively utilize the graph topology for node unlearning. Some methods even compromise the graph's topology, making it challenging to achieve a satisfactory performance-complexity trade-off. To address these issues and achieve efficient unlearning for training node removal in GNNs, we propose three novel node unlearning methods: Class-based Label Replacement, Topology-guided Neighbor Mean Posterior Probability, and Class-consistent Neighbor Node Filtering. Among these methods, Topology-guided Neighbor Mean Posterior Probability and Class-consistent Neighbor Node Filtering effectively leverage the topological features of the graph, resulting in more effective node unlearning. To validate the superiority of our proposed methods in node unlearning, we conducted experiments on three benchmark datasets. The evaluation criteria included model utility, unlearning utility, and unlearning efficiency. The experimental results demonstrate the utility and efficiency of the proposed methods and illustrate their superiority compared to state-of-the-art node unlearning methods. Overall, the proposed methods efficiently remove sensitive training nodes and protect the privacy information of sensitive nodes in GNNs. The findings contribute to enhancing the privacy and security of GNN models and provide valuable insights into the field of node unlearning.
\end{abstract}




\maketitle


\section{Introduction}
A growing number of countries and territories have expressed growing concerns regarding the protection of personal data, prompting the introduction of legislation to address these concerns. Noteworthy examples of such legislation include the General Data Protection Regulation (GDPR) in the European Union \cite{GeneralDataProtection}, the California Consumer Privacy Act (CCPA) in the United States \cite{CaliforniaConsumerPrivacy}. 
These legal frameworks incorporate various provisions, among which the right to be forgotten stands out as both significant and contentious. This particular provision empowers data subjects by granting them the right to request the removal of their personal information from the entities responsible for its maintenance.
Within the realm of machine learning, the right to be forgotten pertains to the obligation of model providers to eliminate any traces of data that individuals have requested to be forgotten. To accomplish this, two methods are available: retraining the model from scratch and machine unlearning \cite{DBLP:journals/csur/XuZZZY24}. However, completely restarting retraining incurs significant computational costs. Therefore, many researchers have studied how to effectively eliminate the impact of sensitive data that they want to remove from the model based on machine unlearning.

Machine unlearning has attracted considerable attention with the emergence of a variety of techniques. The SISA \cite{DBLP:conf/sp/BourtouleCCJTZL21} framework and fine-tuning are two common approaches to machine unlearning nowadays. The core principle of SISA revolves around training individual fragment models by partitioning the training dataset into disconnected fragments. In response to an unlearning request, the model provider retrains the corresponding fragment model. However, instead of mitigating the costs of retraining, the SISA approach chooses to trade storage space for retraining time. When machine unlearning is performed within the SISA framework, portions of the model still need to be retrained. Another widely explored approach to machine learning is parameter fine-tuning of pre-trained models \cite{DBLP:journals/corr/abs-2209-02299}. In this method, machine unlearning is done by fine-tuning to eliminate the effect of the unlearning data on the model parameters, thus achieving the goal of forgetting the data. These two methods are widely used in machine unlearning in various domains \cite{kochno, DBLP:conf/nips/GuptaJNRSW21, DBLP:conf/icml/WuDD20, DBLP:conf/aaai/WuHS22}.



The aforementioned machine unlearning techniques have primarily focused on handling text and image data represented in Euclidean space. However, many real-world datasets, such as social networks \cite{DBLP:conf/aaai/WuLXWC20, DBLP:conf/nips/HamiltonYL17}, citation networks \cite{11072027, 10747296}, and recommender systems \cite{DBLP:conf/kdd/YingHCEHL18, DBLP:conf/www/Fan0LHZTY19}, are structured as graphs. Graph neural networks (GNNs) \cite{DBLP:journals/tnn/ScarselliGTHM09}, a recent class of machine learning models, have emerged as a promising approach for leveraging the rich information embedded in graph data \cite{DBLP:journals/nn/QuanZZXNS25, DBLP:journals/air/GuanZZC24, DBLP:journals/nn/ZhangWHQHG25}.

The current types of unlearning methods in GNNs include node unlearning and link unlearning \cite{DBLP:conf/ccs/Chen000H022}. Node unlearning is focused on efficiently removing specific training nodes from the graph data while safeguarding the privacy of sensitive nodes without compromising the overall model performance. On the other hand, link unlearning involves the removal of specific edges or links between nodes in the graph. Node unlearning holds significant importance due to its ability to protect the privacy of individual nodes in a graph. In various real-world applications, individual nodes within a graph may contain sensitive attributes or data that must be handled with care to adhere to privacy regulations and ethical considerations. By effectively unlearning specific nodes, the model can maintain its utility while preventing potential adversaries from inferring sensitive information about these nodes. This capability of node unlearning makes it a critical tool for privacy preservation in GNNs. Building upon the significance of node unlearning, this paper aims to investigate and explore node unlearning techniques in GNNs. 

SISA and fine-tuning methods are effective for machine unlearning in domains such as images and text. However, when applied to graph data in GNNs, it becomes evident that existing research methods encounter specific challenges in node unlearning:

\begin{itemize}


    \item The SISA method involves randomly dividing the training dataset into disconnected fragments and training individual sub-models for each fragment. This method offers the advantage of retraining only the sub-model associated with the relevant fragments in response to an unlearning request, thereby alleviating the computational resources compared to full retraining. However, it is important to note that directly applying the SISA method to graph data can have undesirable results, compromising the structural integrity of the graph and reducing the utility of GNN models.

    \item The fine-tuning method requires only fine-tuning the parameters of the pre-trained model to accomplish unlearning. Compared to retraining the model from scratch, this method saves a significant amount of time and resources. However, when applied to node unlearning in graph data, it becomes necessary to consider the relationship between the unlearning node and its neighboring nodes. Regrettably, the current methods available do not adequately address the task of node unlearning in the context of graphs. Consequently, the challenge of effectively incorporating neighboring nodes into the node unlearning process remains an open and valuable area of research.
    
\end{itemize}


In this paper, we investigate the parameter fine-tuning method for node unlearning within the context of GNN-based node classification. To tackle the challenges, we propose three distinct methods for node unlearning: Class-based Label Replacement, Topology-guided Neighbor Mean Posterior Probability, and Class-consistent Neighbor Node Filtering. The Class-based Label Replacement method draws inspiration from machine unlearning techniques commonly employed in image and related domains, adapting them for efficient node unlearning in graph data. On the other hand, the Topology-guided Neighbor Mean Posterior Probability and Class-consistent Neighbor Node Filtering methods leverage the topological features inherent in the graph to achieve effective node unlearning.


In the Class-based Label Replacement method, node unlearning can be effectively achieved even in the absence of node topology features. This technique involves substituting the target class of the unlearning node with the mean posterior probability of that same class. The method comprises several key steps: firstly, the mean posterior probability of each class in the testing set is computed. Subsequently, the target class of the unlearning node is replaced with the calculated mean posterior probability corresponding to the same class. Finally, the model is fine-tuned using the adjusted target class to mitigate the influence of the unlearning node on the GNN model.


Topology-guided Neighbor Mean Posterior Probability and Class-consistent Neighbor Node Filtering leverage the topological structure features of the graph. Both methods utilize the graph's topology to identify the neighbors of the unlearning nodes and combine their posterior probabilities to facilitate machine unlearning. The Topology-guided Neighbor Mean Posterior Probability method identifies each neighboring node of the unlearning nodes and replaces the target class of the unlearning nodes with the mean posterior probability calculated from the neighboring nodes. On the other hand, the Class-consistent Neighbor Node Filtering method identifies the neighboring nodes that belong to the same class as the unlearning node and are not present in the training set. It then replaces the target class of the unlearning node with the mean posterior probabilities of these selected neighboring nodes.



We conducted experiments on widely recognized citation network datasets, namely Cora, Citeseer, and Pubmed \cite{DBLP:conf/iclr/KipfW17}, to evaluate the effectiveness of our proposed method. Additionally, we employed membership inference attacks as a validation method to assess the efficacy of our methods in reducing the impact of unlearning nodes. The experimental results demonstrate the success of our three proposed unlearning methods in effectively removing nodes from the model while preserving the model's performance. 

In summary, our study makes the following contributions:
\begin{itemize}

    \item We introduce three distinct methods for node unlearning in GNN-based node classification tasks.

    \item Our proposed method exhibits high efficiency in node unlearning while preserving the model's performance.

    \item We validate the utility of node unlearning using a membership inference attack, confirming the effectiveness of our proposed node unlearning approach.


\end{itemize}

\section{Preliminaries}
\subsection{Notations}

We define a graph as $G(X, Y, A)$, where $X$ represents the node features, $Y$ represents the corresponding labels, and $\left|G\right|$ denotes the total number of nodes in the graph. The nodes $v$ and $u$ refer to specific nodes within the graph, while $N(v)$ denotes the set of neighboring nodes of node $v$. The loss function is represented by $\mathcal{L}$, and the training objective is denoted as $\mathcal{J}$. The weight matrix is represented by $W$, and $\theta$ represents the model parameters. The node embedding features are denoted as $Z$, which capture the node representations at each layer of the graph neural network (GNN). Additionally, $I_N$ denotes the identity matrix, and $C$ represents the node class. The notations used in this paper are summarized in Table \ref{Notations}.

\begin{table}[tbp]
  \centering
  \caption{Summary of notations.}
    \begin{tabular}{c|c}
    \toprule
    \textbf{Notation} & \textbf{Description} \\
    \midrule
    $G(X, Y, A)$ & Graph \\
    $X$     & Node feature \\
    $Y$     & Node Label \\
    $A$     & Adjacency matrix \\
    $\left|G\right|$   & Number of nodes \\
    $u$,$v$     & Node in graph \\
    $\mathcal{N}(v)$  & The neighboring nodes of node \$v\$ \\
    $\mathcal{L}$    & Loss function\\
    $\mathcal{J}$     &  Training objective \\
    $W$     & Weight matrix \\
    $\theta$ & Model parameters \\
    $H$     & Hidden state of node feature \\
    $H$     & Posterior probability \\
    $I_N$    & Identity matrix \\
    $C$     & Classes \\
    \bottomrule
    \end{tabular}%
  \label{Notations}%
\end{table}%

\subsection{Machine Unlearning}

With the enforcement of privacy protection regulations, it has become crucial for users to have control over their data and be able to request its removal from a network. Similarly, in machine learning, model providers are expected to remove revoked data from the training set while ensuring minimal impact on the model's performance. In recent years, machine unlearning has emerged as a prominent research area in AI security and privacy \cite{DBLP:conf/sp/CaoY15}. Machine unlearning refers to the process of modifying a trained machine learning model to forget or remove specific data points from the model's training set. This process enables the model to quickly forget data, addressing privacy concerns and complying with data protection regulations. In the following sections, we provide an overview of the basic forgotten data methods and the mainstream techniques used in machine unlearning.


\textbf{Retraining from Scratch.} Retraining a machine learning model from scratch is the simplest way to implement forgotten data. In this method, we first remove the data that we want to forget from the training set and then train a new model from scratch using the remaining dataset. Retraining from scratch is an effective technique for forgetting data for privacy protection as it completely removes the knowledge of the forgotten data and is easy to implement. However, this approach can be computationally expensive, especially when dealing with complex models and large training datasets.


\textbf{SISA}. SISA is an ensemble-based machine unlearning method proposed by Bourtoule et al. \cite{DBLP:conf/sp/BourtouleCCJTZL21}, has gained significant popularity in recent years. SISA is designed to work with various types of machine learning models and offers a practical approach to unlearning. It achieves this by training a single fragmented model, where the training dataset is randomly partitioned into disconnected subsets. The final prediction is obtained by combining the individual predictions from these fragmented models. When a model provider receives an unlearning request, only the relevant fragment models need to be retrained, reducing the overall cost of unlearning. However, it is important to note that if the unlearning data is distributed across multiple fragmented sections, a considerable number of models may need to be retrained. In the worst-case scenario, every model would require retraining.


\textbf{Fine-tuning.} Another common method in machine unlearning is to reduce the influence of unlearning data on a trained model by fine-tuning its parameters. During the initial training phase of the model, the unlearning data contributes to the learning process along with the normal training data. However, in the subsequent unlearning process, the model owner can fine-tune the model parameters by adjusting the loss function associated with the unlearning data. This adjustment aims to minimize the impact of the unlearning data on the model's performance. By carefully selecting an appropriate substitution loss for the unlearning data, this method enables efficient machine unlearning with relatively low computational cost.

\subsection{Graph Neural Networks (GNNs)}

The graph is a non-fixed, irregular data structure that poses challenges for conventional neural network architectures such as Convolutional Neural Networks (CNNs) \cite{DBLP:conf/nips/KrizhevskySH12} and Recurrent Neural Networks (RNNs) \cite{DBLP:conf/interspeech/MikolovKBCK10}. The fixed formats of these networks limit their direct applicability to graph-structured data. To address this issue, researchers have proposed Graph Neural Networks (GNNs) \cite{DBLP:journals/tnn/ScarselliGTHM09}, which extend existing neural network methods to effectively handle graph data.


The topology and node features of graphs can be effectively captured and represented by GNNs. GNNs employ a unique message passing mechanism that enables the incorporation of both topological information and node features in graph data. To update the node representations at each layer, GNNs aggregate information from neighboring nodes, allowing for a more comprehensive understanding of the graph structure. The updated node representation in the $k$-th layer of the GNN can be mathematically expressed using the following formula:

\begin{equation}
    \begin{gathered}
        \mathrm{h}_{v}^{(k)}=\operatorname{COMBINE}^{(k)}\left(\mathrm{h}_{v}^{(k-1)}, \mathrm{m}_{v}^{(k)}\right)\\
        \mathrm{m}_{v}^{(k)} = \text {AGG}^{(k-1)}\left(\left\{\mathrm{h}_{u}^{(k-1)}: \forall u \in \mathcal{N}(v) \cup v\right\}\right)
    \end{gathered}
\end{equation}
where $\mathcal{N}(v)$ represents the neighboring nodes of node $v$, and $u$ denotes a node within the set of neighboring nodes. ${h}_{v}^{(k)}$ refers to the hidden state of node in the $k$-th layer, while ${h}_{v}^{(0)}$ denotes the input feature of node $v$. The function \text {AGG} is responsible for aggregating the information from the neighboring nodes.


GNNs have demonstrated their effectiveness in a wide range of graph-related tasks, including node-level tasks \cite{11088144}, link-level tasks \cite{DBLP:conf/uss/HeJ0G021}, and graph-level tasks \cite{DBLP:conf/ccs/MuWL0XL21}. To cater to the diverse requirements of these tasks, several variations of GNNs have been proposed, with graph convolutional neural networks (GCN) \cite{DBLP:conf/iclr/KipfW17} being the most prevalent. In this study, we specifically focus on research related to GCN-based node unlearning.

\section{Problem Definition}
\subsection{Graph Unlearning Definition}

When the model owner receives a request to delete specific data, two methods can be employed for data deletion: retrain and unlearning. The retrain method involves training the model from scratch using a training set that does not include the deleted data. However, this approach can be time-consuming and resource-intensive, especially for large models, resulting in unnecessary waste. On the other hand, unlearning is a more efficient method that removes the data from the trained model while preserving the model's performance.

\begin{figure}[ht]
\centering
\includegraphics[scale=0.13]{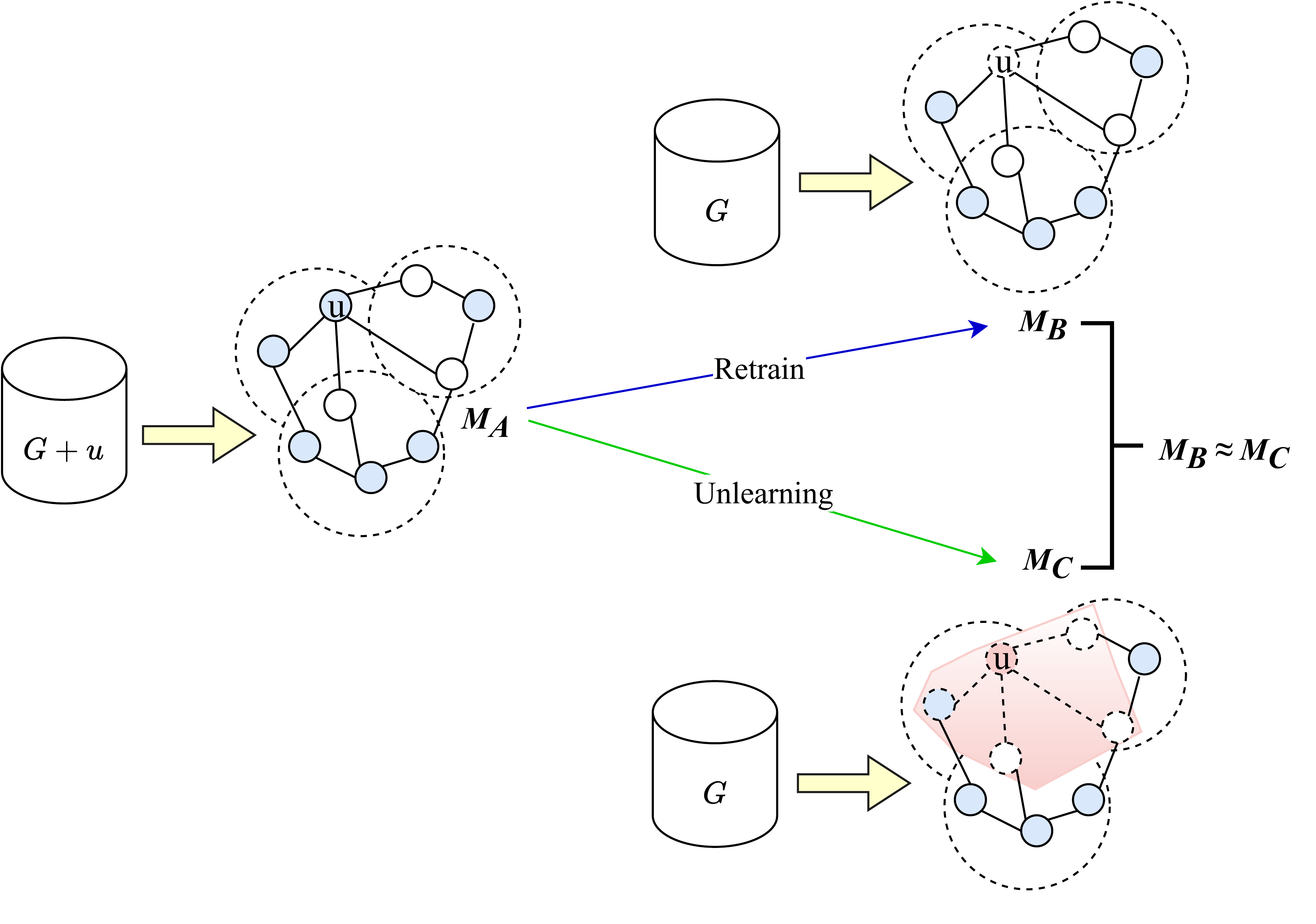}
\caption{Removing data from a trained model. $G+{u}$ denotes that the training dataset contains unlearning nodes $u$, and $G$ denotes that the training dataset does not contain unlearning nodes $u$.}
\label{Defining}
\end{figure}

For an existing model \textbf{\emph{$M_A$}} that has already been trained, when it is necessary to remove the forgetting node $u$ (unlearning node), the retrain method retrains the model using the training set without the node $u$ to obtain the model \textbf{\emph{$M_B$}}, as shown in Fig. \ref{Defining}. In contrast, unlearning modifies the existing model \textbf{\emph{$M_A$}} to quickly accomplish forgetting the node $u$, thus obtaining the model \textbf{\emph{$M_C$}} without the node $u$. The goal of unlearning is to quickly forget the specified data and obtain a model \textbf{\emph{$M_C$}} with similar performance to the model \textbf{\emph{$M_B$}} obtained through retrain, i.e., \textbf{\emph{$M_C$}} $\approx$ \textbf{\emph{$M_B$}}.

\begin{figure}[ht]
\centering
\includegraphics[scale=0.1]{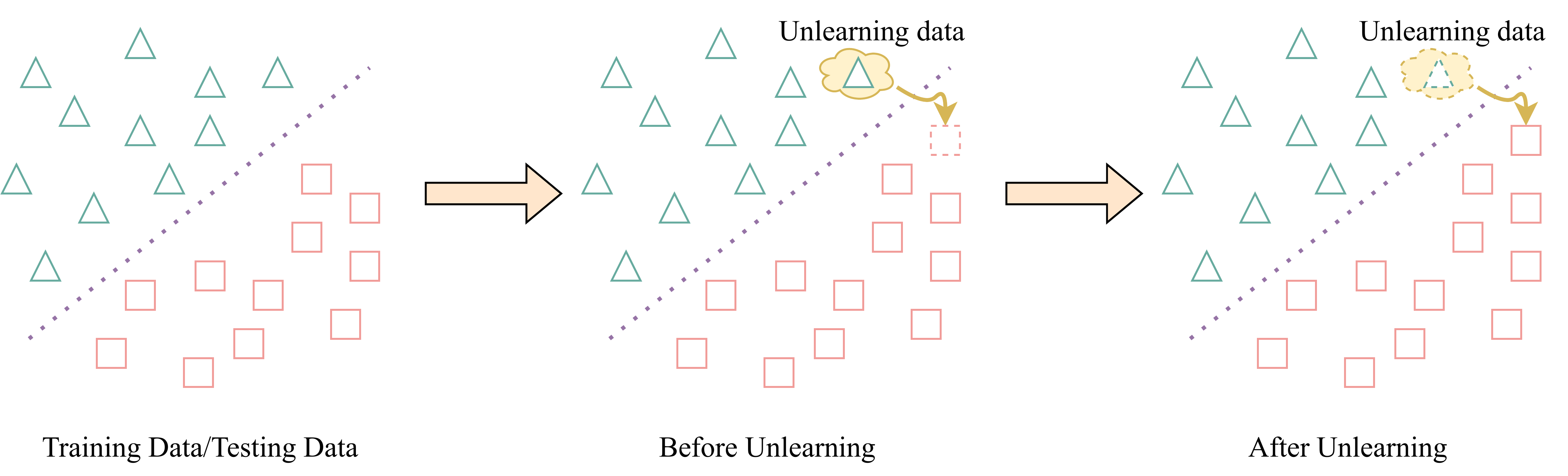}
\caption{Data distribution of training and testing sets. The $\triangle$ represents the training data, while $\square$ represents the testing data. The goal of unlearning is to remove specific data instances from the training set while maintaining the ability to identify those instances as if they were part of the testing set.}
\label{train and test}
\end{figure}

In addition, it is worth mentioning that unlearning should ensure that the model recognizes the unlearning data with the same accuracy as the data in the testing set. In other words, unlearning is not misrecognizing the unlearning data but recognizing it as data that the model has not seen before, i.e., the testing data. Therefore, as shown in Fig. \ref{train and test}, the process of unlearning can be viewed as removing the unlearning data from the training set so that the model forgets that data, while maintaining the performance on the unlearning data.

\subsection{Base Model}

\begin{figure*}[t]
\centering
\includegraphics[scale=0.18]{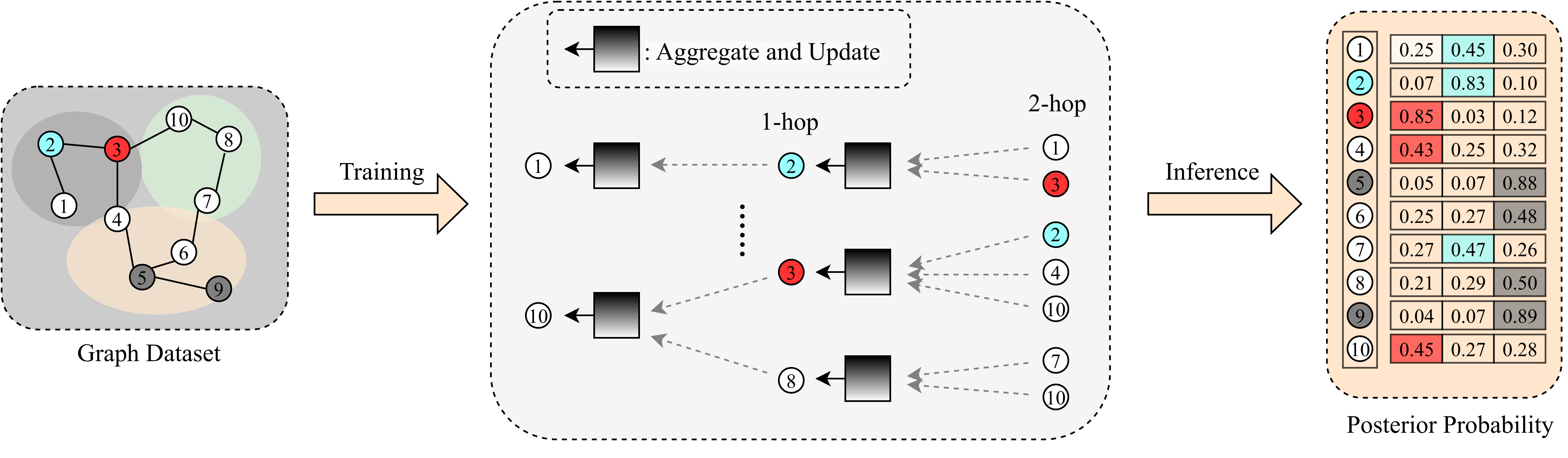}
\caption{The general design pipeline for a GNN model. The dataset contains 10 nodes, with different colors representing different classes of nodes in the training set. The blank nodes correspond to unlabeled nodes, which are equivalent to nodes in the testing set.}
\label{GCN prediction}
\end{figure*}

In this study, we focus on node classification, and specifically, we employ a two-layer GCN for this task. To prepare the data for the GCN model, we perform a pre-processing step known as $\hat{A}=\tilde{D}^{-\frac{1}{2}} \tilde{A} \tilde{D}^{-\frac{1}{2}}$, where $\tilde{A}=A+I_N$ represents the adjacency matrix of the undirected graph with added self-connections. Here, $I_N$ denotes the identity matrix, and $\tilde{D}_{i i}=\sum_j \tilde{A}_{i j}$ represents the diagonal matrix with row sums of $\tilde{A}$. Subsequently, we employ a straightforward form for our forward model:
\begin{equation}
    \begin{gathered}
        Z=f(X, A)=\operatorname{softmax}\left(\hat{A} \operatorname{ReLU}\left(\hat{A} X W^{(0)}\right) W^{(1)}\right)
    \end{gathered}
\end{equation}
where $W^{(0)} \in \mathbb{R}^{C \times H}$ represents the input-to-hidden weight matrix of the hidden layer with $H$ feature maps. $W^{(1)} \in \mathbb{R}^{H \times F}$ denotes the weight matrix from the hidden layer to the output layer. To obtain the node-wise predictions, we apply the softmax activation function row-wise, defined as $\operatorname{softmax}\left(x_i\right)=\frac{1}{\mathcal{Z}} \exp \left(x_i\right)$, where $\mathcal{Z}=\sum_i \exp \left(x_i\right)$ represents the normalization constant. As depicted in Fig. \ref{GCN prediction}, we input the graph data into the GCN model, which utilizes message-passing techniques such as aggregation and updating to generate the posterior probability $\mathcal{Z}$ for each node.

The weights $W^{(0)}$ and $W^{(1)}$ of the neural network are optimized through the training process using a loss function. Subsequently, we evaluate the loss for all labeled samples in the context of node multi-class classification, which can be expressed as follows:
\begin{equation}
    \begin{gathered}
        \mathcal{L}=-\sum_{l \in \mathcal{Y}_L} \sum_{f=1}^F Y_{l f} \ln Z_{l f}
    \end{gathered}
\end{equation}
where $\mathcal{Y}_L$ is the collection of labeled node indices.


In GCN, the loss is computed on all labeled data, and the model is iteratively trained by applying gradient updates based on this computed loss. As a result, the trained model becomes highly correlated with the labeled data used during training. However, the unlearning method proposed in this study aims to reduce the correlation between the model and specific data instances. This approach aligns with the objectives of data forgetting and privacy protection, as it ensures that sensitive information is no longer retained within the model.

In the following section, we introduce the node unlearning method proposed in this paper.

\subsection{Node Unlearning}

In this paper, our research focuses on the task of node unlearning. The entire graph dataset is denoted as $G(X, Y, A)$. Furthermore, we use $G_L$ to represent the set of labeled nodes in the graph.

Initially, we obtain the model \textbf{\emph{$M_A$}} as shown in Fig. \ref{Defining} through training on the labeled set $G_L$. The parameters of the model \textbf{\emph{$M_A$}} are denoted by $\theta_\text{org}$. It is assumed that $\theta_\text{org}$ represents the optimal solution of the dataset $G_L$, given by:
\begin{equation}
    \begin{gathered}
        \theta_{\text {org }}=\underset{\theta}{\operatorname{argmin}} \mathcal{J}_{\text {org }}(\theta)=\underset{\theta}{\operatorname{argmin}} \frac{1}{\left|G_L\right|} \sum_{i=1}^{\left|G_L\right|} \mathcal{L}\left(\mathbf{x}_i, y_i, A, \theta\right)
    \end{gathered}
\end{equation}
where ${\left|G_L\right|}$ denotes the number of labeled nodes. 



When a request for node unlearning is received, we define the set of forgetting nodes (unlearning nodes) as $G_u$, and the set of nodes after removing the forgetting node is denoted by $G_p$, where $G_p = G_L \backslash G_u$. 

In the retraining method, the \textbf{\emph{$M_B$}} model is retrained using $G_p$ with parameters denoted by $\theta_{\text{retrain}}$ as follows:
\begin{equation}
    \begin{gathered}
        \theta_{\text {retrain }}=\underset{\theta}{\operatorname{argmin}} \mathcal{J}_{\text {retrain }}(\theta)=\underset{\theta}{\operatorname{argmin}} \frac{1}{\left|G_p\right|} \sum_{i=1}^{\left|G_p\right|} \mathcal{L}\left(\mathbf{x}_i, y_i, A, \theta\right)
    \end{gathered}
\end{equation}

In the unlearning process, we derive the \textbf{\emph{$M_C$}} parameters denoted as $\theta_{\text{unlearning}}$ from the original parameters $\theta_{\text{org}}$, where $\theta_{\text {unlearning }} \approx \theta_{\text {retrain }}$, and $\theta_{\text {org }} - \theta_{\text {unlearning }} = $


\begin{equation}\label{loss}
    \begin{gathered}
        \underset{\theta}{\operatorname{argmin}} \underbrace{\frac{1}{\left|G_L\right|} \sum_{i=1}^{\left|G_L\right|} \mathcal{L}\left(\mathbf{x}_i, y_i, A, \theta\right)}_{\mathcal{J}_{\text {org }}(\theta)}+\underbrace{\frac{1}{\left|G_u\right|} \sum_{j=1}^{\left|G_u\right|}  \mathcal{L}\left(\mathbf{x}_j, y_j, A, \theta\right)}_{\mathcal{J}_{\text {unlearning }}(\theta)}
    \end{gathered}
\end{equation}



In comparison to $\theta_{\text{retrain}}$ obtained through retraining, $\theta_{\text{unlearning}}$ can be obtained more quickly while still maintaining the same accuracy rate.

In the task of node unlearning, it is crucial to consider not only the individual node being unlearned but also its impact on other nodes, particularly its neighboring nodes, as illustrated by the shaded part in model $M_C$ in Fig. \ref{Defining}. Thus, a key focus of this paper is to investigate the integration of neighboring nodes design ${\mathcal{J}_{\text {unlearning }}(\theta)}$ to enhance the modification of the model for more effective node unlearning.

\section{Propose Methodology}

In this paper, we introduce three distinct yet synergistic approaches to node unlearning: Class-based Label Replacement, Topology-guided Neighbor Mean Posterior Probability, and Class-consistent Neighbor Node Filtering. The Class-based Label Replacement method enables node unlearning while considering the lost topological structure feature of the model owner. The Topology-guided Neighbor Mean Posterior Probability method contributes to partial unlearning of nodes while maximizing model utility. On the other hand, the Class-consistent Neighbor Node Filtering method achieves effective node unlearning while sustaining high model utility. Notably, the Class-consistent Neighbor Node Filtering method emerges as the most efficient approach in terms of overall performance. These methods are designed with the common goal of preserving privacy while minimizing the negative impact on model performance. In the following sections, we will discuss the similarities and differences between each method. 


\subsection{Class-based Label Replacement (CLR)}




Drawing on machine unlearning research in different fields, we propose a novel approach to graph node unlearning that we call Class-based Label Replacement (CLR). This method involves replacing the label of the unlearning node with the mean posterior probability of the same class in the testing set. To begin, we calculate the mean posterior probability ($\overline{Z}$) for each class in the testing set. Next, we update the target label of the unlearning node by assigning it the mean posterior probabilities of the same classes in the testing set.

\begin{figure}[h]
\centering
\includegraphics[scale=0.16]{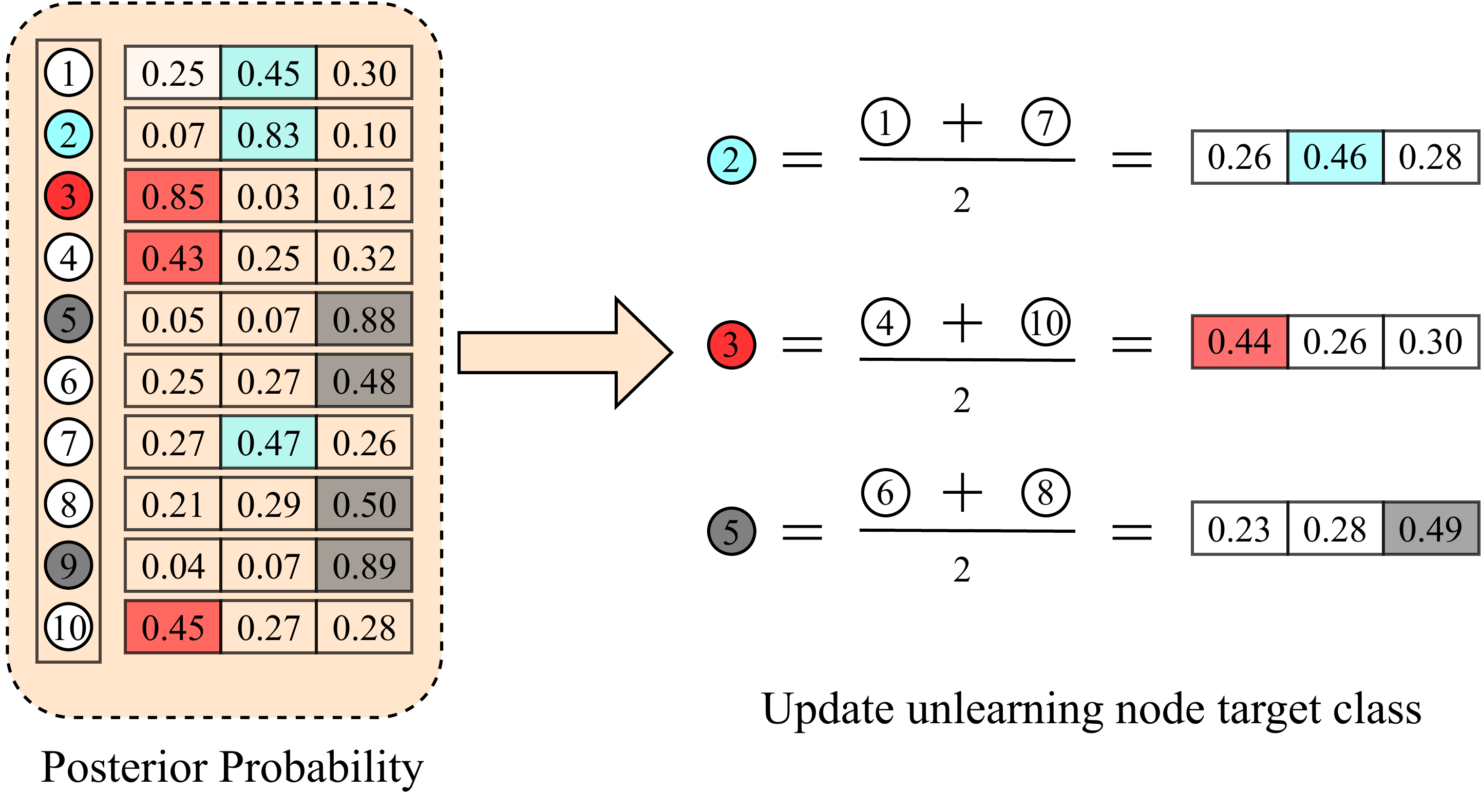}
\caption{Class-based Mean Posterior Probability Label Replacement for Unlearning.}
\label{mean}
\end{figure}
As shown in Fig. \ref{mean}, node $2$ is one of the unlearning nodes. We compute the mean posterior probability of unlabeled nodes $1$ and $7$, which are of the same class as unlearning node $2$, and then replace the target class of node $2$ with the obtained mean posterior probability. By updating the target class ($Y'$), we can effectively accomplish unlearning. The goal of unlearning can be formulated as follows:
\begin{equation}\label{unlearning train}
    \begin{gathered}
        {\mathcal{J}_{\text {unlearning }} } = \frac{1}{\left|G_u\right|} \sum_{j=1}^{\left|G_u\right|}  \mathcal{L}\left(\mathbf{x}_j, y^{\prime}_j, A,  \theta\right)
    \end{gathered}
\end{equation}


Algorithm \ref{algorithm mean} outlines the workflow of the proposed Class-based Label Replacement method. This algorithm takes several inputs, including node features ($X$), adjacency matrix ($A$), training node index ($idx_\text{train}$), unlearning node index ($idx_\text{unlearning}$), testing node index ($idx_\text{test}$), node posterior probabilities ($Z$), node labels ($Y$), and original model parameters ($\theta_{\text {org }}$). The objective of the algorithm is to obtain the updated model parameters ($\theta_{\text {unlearning }}$) through the training process. The algorithm operates in three distinct steps, as described below:
\begin{itemize}


    \item \textbf{Step 1: Calculation of Mean Posterior Probability (Line \ref{algorithm1 step1}).} In this step, we compute the mean posterior probability of each class in the testing set using the provided testing set index ($idx_\text{test}$), node labels ($Y$), and node posterior probabilities ($Z$). Specifically, for a given class $C$, the mean posterior probability ($\overline{Z'_{\text {c }} }$) is computed according to the following procedure:

    \begin{equation}\label{mean class}
        \begin{gathered}
            \overline{Z'_{\text {c }} } = \frac{1}{\left|Y_{i} = c\right|} \sum_{i}^{\left|Y_{i} = c\right|}  Z_i, i \in idx_{test}, Y_{i} = c
        \end{gathered}
    \end{equation}


    \item \textbf{Step 2: Modification of Unlearning Nodes' Labels (Line \ref{algorithm1 step21} - Line \ref{algorithm1 step22}).} In this step, we focus on modifying the labels of the unlearning nodes while keeping the labels of the remaining nodes unchanged. Specifically, we update the label ($Y_v$) of an unlearning node $v$ to the mean posterior probability of its corresponding class ($\overline{Z'_{\text {c }} }$). The formula used for updating the label of the unlearning node $v$ can be illustrated by the following example:
    
    \begin{equation}\label{modify}
        \begin{gathered}
             Y^{\prime}_v= \overline{Z'_{\text {c }} }, Y_v = c, v \in idx_{unlearning}
        \end{gathered}
    \end{equation}


    \item \textbf{Step 3: Model Training (Line \ref{algorithm1 step31} - Line \ref{algorithm1 step32}).} In this step, we proceed with training the model using the updated labels obtained in Step 2. The target label utilized for calculating the model's loss is the label that has been modified during the previous step, as shown in Eq. \ref{unlearning train}. The model's parameters are then iteratively updated based on the calculated losses, ensuring the refinement of the model's performance.

\end{itemize}

\begin{algorithm}[t]
\caption{Unlearning with Class-based Label Replacement.}\label{algorithm mean} 
\hspace*{0.02in} {\bf Input:} 
$X$, $A$, $idx_\text{train}$, $idx_\text{test}$, $idx_\text{unlearning}$, $Z$, $Y$, and $\theta_{\text {org }}$\\
\hspace*{0.02in} {\bf Output:} 
$\theta_{\text {unlearning }}$
\begin{algorithmic}[1]
\State Calculate each class's mean posterior probability $\overline{Z}$ by Eq.(\ref{mean class}). \label{algorithm1 step1}

\For{$i\le n$}\label{algorithm1 step21} 
    \If{$i \in idx_\text{unlearning}$} 
        \State Modify the target class of node $Y_i$ to $Y^{\prime}_i$ by Eq.(\ref{modify}).
    \Else
        \State No modification in target class, $Y^{\prime}_i = Y_i$.
    \EndIf \label{algorithm1 step22}
\EndFor
\Repeat \label{algorithm1 step31}
    \State Obtain $\theta_{\text {unlearning }}$ with $G(X, A, Y^{\prime}_i)$, $\theta_{\text {org }}$ and Eq.(\ref{loss}). 
\Until convergence \label{algorithm1 step32}\\ 
\Return $\theta_{\text {unlearning }}$
\end{algorithmic}
\end{algorithm}

In contrast to traditional image and text data, graph data possesses distinct characteristics, including both node features and unique topological structure features. While the Class-based Label Replacement method utilizes the posterior probability of the testing set to modify the target class, it overlooks the crucial topological structure features inherent in graph data. Neglecting the graph topology during the unlearning process can yield suboptimal outcomes. The decisions made during unlearning fail to account for the relationships and dependencies among nodes within the graph, subsequently compromising the effectiveness of the unlearning. Consequently, the CLR method may retain residual influence or connections of the unlearning nodes that should have been severed based on removing the unlearning nodes.

\subsection{Topology-guided Neighbor Mean Posterior Probability (TNMPP)}

Taking into account the inherent topological structure of the graph, we propose a novel unlearning method called Topology-guided Neighbor Mean Posterior Probability (TNMPP). This method facilitates node unlearning by leveraging the posterior probability of neighboring nodes. Initially, the method identifies the neighboring nodes of the unlearning nodes based on the graph's topology. Subsequently, it calculates the mean posterior probabilities of these neighbor nodes. Finally, the target class of the unlearning node is modified by the mean posterior probability of the neighboring nodes.

\begin{figure}[ht]
\centering
\includegraphics[scale=0.13]{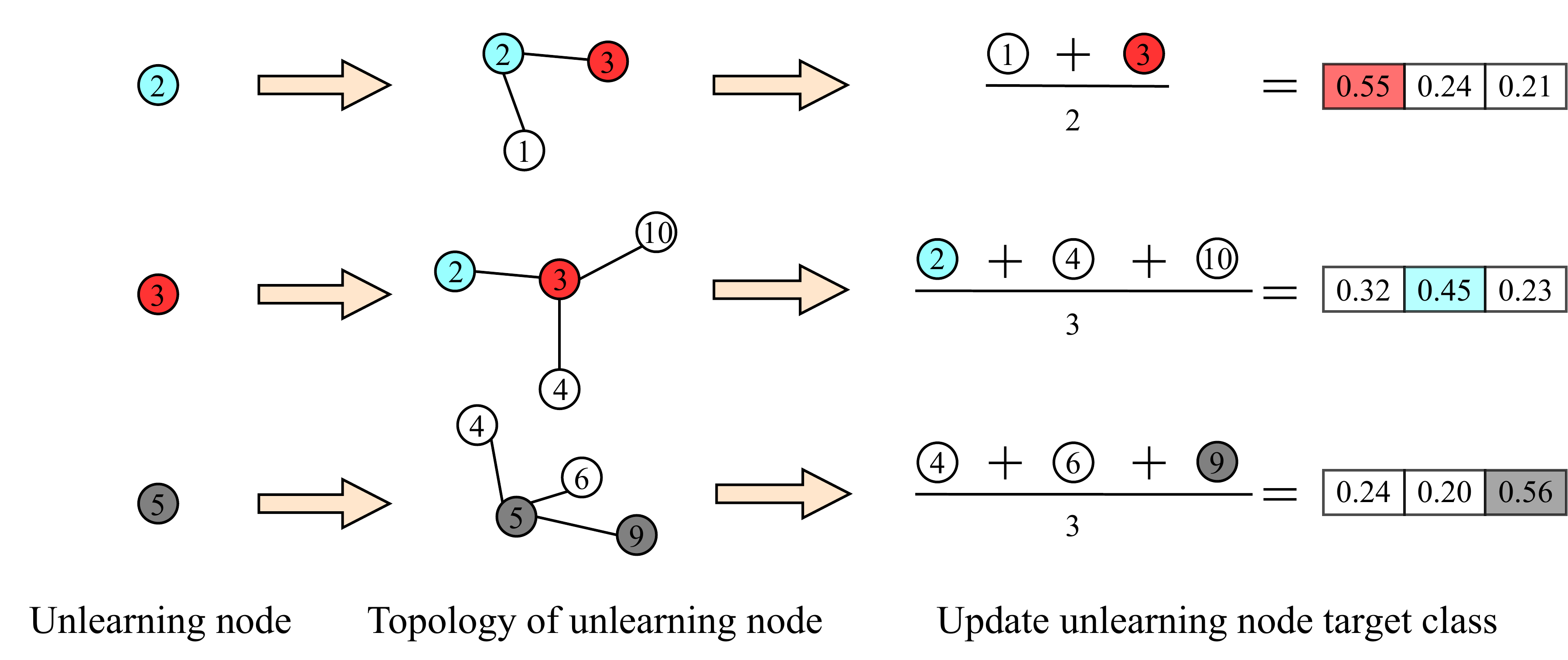}
\caption{Topology-guided Neighbor Mean Posterior Probability for Unlearning.}
\label{node}
\end{figure}

As depicted in Fig. \ref{node}, the unlearning node $2$ serves as an illustration of this process, where its target class is adjusted based on the information from neighboring nodes $1$ and $3$. More precisely, the unlearning of node $2$ entails modifying its target class through the computation of the mean posterior probabilities obtained from the neighboring nodes $1$ and $3$.

Algorithm \ref{algorithm same neighbor} outlines the workflow for Topology-guided Neighbor Mean Posterior Probability. The algorithm accepts several inputs, including node features ($X$), adjacency matrix ($A$), training node index ($idx_\text{train}$), unlearning node index ($idx_\text{unlearning}$), testing node index ($idx_\text{test}$), node posterior probabilities ($Z$), node labels ($Y$), and original model parameters ($\theta_{\text {org }}$). Upon training the model, we obtain the updated parameters ($\theta_{\text {unlearning }}$). The algorithm can be divided into four distinct steps, as described below:

\begin{itemize}


    \item \textbf{Step 1: Neighbor Node Identification (Line \ref{algorithm2 step1}).} The first step involves identifying the neighboring nodes ($\mathcal{N}(v)$) of the unlearning node ($v$) using the adjacency matrix ($A$). This matrix enables us to determine the set of nodes that are directly connected to the unlearning node.
    

    \item \textbf{Step 2: Computation of Mean Posterior Probability for Neighbor Nodes (Line \ref{algorithm2 step2}).} In this step, we calculate the mean posterior probability ($\overline{Z'_{\text {v }} }$) of the neighbor nodes ($\mathcal{N}(v)$) that are connected to the unlearning node ($v$). The calculation process is depicted as follows:

    \begin{equation}\label{mean neighbor}
        \begin{gathered}
            \overline{Z'_{\text {v }} } = \frac{1}{\left|\mathcal{N}(v)\right|} \sum_{u}^{\left|\mathcal{N}(v)\right|}  Z_u, \forall u \in \mathcal{N}(v)
        \end{gathered}
    \end{equation}


    \item \textbf{Step 3: Adjustment of Unlearning Nodes' Labels (Line \ref{algorithm2 step3}).} In this step, we update the label ($Y_v$) of the unlearning node ($v$) by assigning it the mean posterior probability ($\overline{Z'_{\text {v }} }$) computed from its neighboring nodes. The formula for updating the label of node $v$ can be expressed as follows:
    \begin{equation}\label{modify neighbor}
        \begin{gathered}
             Y^{\prime}_v= \overline{Z'_{\text {v }} }, v \in idx_{unlearning}
        \end{gathered}
    \end{equation}


    \item \textbf{Step 4: Model Training (Line \ref{algorithm2 step41} - Line \ref{algorithm2 step42}).} In this step, we calculate the loss using the updated labels obtained from Step 3. Subsequently, we train the model using this loss and update the model parameters accordingly.

\end{itemize}

\begin{algorithm}[t]
\caption{Unlearning with Topology-guided Neighbor Mean Posterior Probability.}\label{algorithm same neighbor} 
\hspace*{0.02in} {\bf Input:} 
$X$, $A$, $idx_\text{train}$, $idx_\text{test}$, $idx_\text{unlearning}$, $Z$, $Y$, and $\theta_{\text {org }}$\\
\hspace*{0.02in} {\bf Output:} 
$\theta_{\text {unlearning }}$
\begin{algorithmic}[1]

\For{$i\le n$} 
    \If{$i \in idx_\text{unlearning}$} 
        \State Finding neighbor nodes ($\mathcal{N}(v)$) of unlearning node ($v$) according to $A$. \label{algorithm2 step1}
        \State Calculate the mean posterior probability of the neighbor nodes ($\overline{Z'_{\text {v }} }$) by Eq.(\ref{mean neighbor}). \label{algorithm2 step2}
        \State Modify the target class of node $v$ to $\overline{Z'_{\text {v }} }$ by Eq.(\ref{modify neighbor}). \label{algorithm2 step3}
        
    \Else
        \State No modification in target class, $Y^{\prime}_v = Y_v$.
    \EndIf
\EndFor
\Repeat \label{algorithm2 step41}
    \State Obtain $\theta_{\text {unlearning }}$ with $G(X, A, Y^{\prime}_i)$, $\theta_{\text {org }}$ and Eq.(\ref{loss}).
\Until convergence \label{algorithm2 step42}\\ 
\Return $\theta_{\text {unlearning }}$
\end{algorithmic}
\end{algorithm}

There are various types of links in the graph, including both meaningful relationships that capture the inherent data structure and unrelated links that may be generated by noise or random connections. In the Topology-guided Neighbor Mean Posterior Probability method, connections of unrelated nodes can significantly influence Step 2 of the unlearning process. For instance, in Fig. \ref{GCN prediction}, node $2$ initially belongs to class \emph{Blue}, but the mean posterior probabilities of its neighboring nodes are shown to be class \emph{Red}. Moreover, neighboring nodes might also contain training set nodes, which can affect the efficacy of unlearning and impede the removal of unlearning nodes from the training set.

\subsection{Class-consistent Neighbor Node Filtering (CNNF)}


To mitigate the influence of unrelated nodes on the unlearning process, we propose a novel approach called Class-consistent Neighbor Node Filtering (CNNF). This method focuses on considering only the neighboring nodes that are of the same class as the unlearning node and are not part of the training set. Initially, we identify all the neighbor nodes connected to the unlearning node. Subsequently, we remove the neighbor nodes that belong to the training set or have different classes from the unlearning node. Finally, we compute the average posterior probability using the remaining nodes, which are the neighbor nodes with the same class in the testing set.

\begin{figure}[ht]
\centering
\includegraphics[scale=0.14]{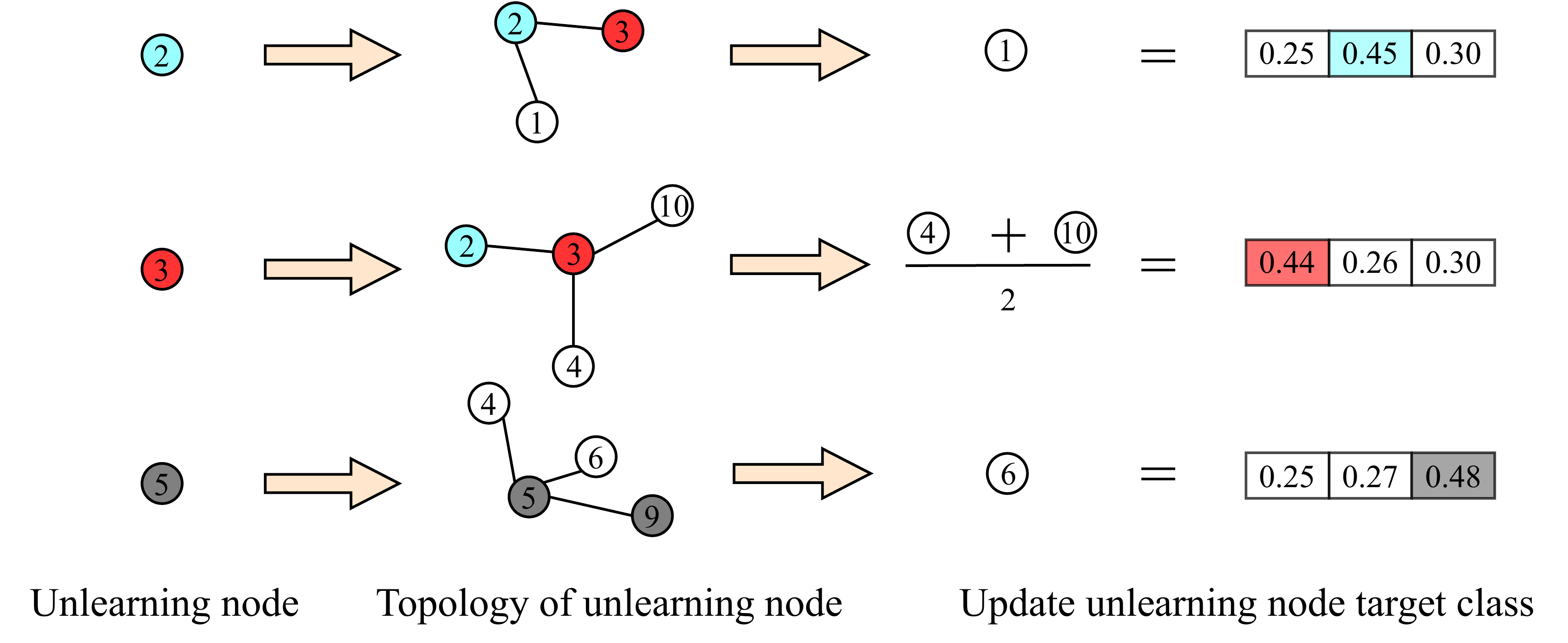}
\caption{Class-consistent Neighbor Node Filtering for Unlearning.}
\label{node same}
\end{figure}

As illustrated in Fig. \ref{node same}, let's consider the unlearning node 3, which has three neighboring nodes: $2$, $4$, and $10$. Among these, nodes $4$ and $10$ belong to the same class as the unlearning node $3$ and are present in the testing set. Node $2$ is neither in the same category as node $3$ and is also a training set node. Therefore, we update the target label of unlearning node $3$ based on the mean posterior probabilities only derived from nodes $4$ and $10$.

\begin{algorithm}[t]
\caption{Unlearning with Class-consistent Neighbor Node Filtering.} \label{algorithm same mean neighbor} 
\hspace*{0.02in} {\bf Input:} 
$X$, $A$, $idx_\text{train}$, $idx_\text{test}$, $idx_\text{unlearning}$, $Z$, $Y$, and $\theta_{\text {org }}$\\
\hspace*{0.02in} {\bf Output:} 
$\theta_{\text {unlearning }}$
\begin{algorithmic}[1]
\State Calculate each class's mean posterior probability $\overline{Z}$ by Eq.(\ref{mean class}).

\For{$i\le n$} 
    \If{$i \in idx_\text{unlearning}$} 
        \State Finding neighbor nodes ($\mathcal{N}(v)$) of unlearning node ($v$) according to $A$. \label{algorithm3 step1}
        \State Selecting neighbor nodes to create a new set of neighbor nodes ($\mathcal{N}(v_u)$) by \ref{selet neighbor}. \label{algorithm3 step2}
        \State Calculate the new set of neighbor nodes' mean posterior probability ($\overline{Z'_{\text {v }} }$) by Eq.(\ref{same mean neighbor}). \label{algorithm3 step3}
        \If{${\left|\mathcal{N}(v_u)\right|} > 0$}\label{algorithm3 step41} 
            \State Modify the target class of node $v$ to $\overline{Z'_{\text {v }} }$.
        \ElsIf{${\left|\mathcal{N}(v_u)\right|} = 0$}
            \State Modify the target class of node $v$ to $\overline{Z'_{\text {c }} }, Y_v = c$.
        \EndIf    \label{algorithm3 step42} 
    \Else
        \State No modification in target class, $Y^{\prime}_v = Y_v$.
    \EndIf
\EndFor
\Repeat \label{algorithm3 step51}
    \State Obtain $\theta_{\text {unlearning }}$ with $G(X, A, Y^{\prime}_i)$, $\theta_{\text {org }}$ and Eq.(\ref{loss}).
\Until convergence \label{algorithm3 step52}\\ 
\Return $\theta_{\text {unlearning }}$
\end{algorithmic}
\end{algorithm}

Algorithm \ref{algorithm same mean neighbor} presents the workflow for Class-consistent Neighbor Node Filtering. It takes various inputs, including node features ($X$), adjacency matrix ($A$), training node index ($idx_\text{train}$), unlearning node index ($idx_\text{unlearning}$), testing node index ($idx_\text{test}$), node posterior probabilities ($Z$), node labels ($Y$), and original model parameters ($\theta_{\text {org }}$). After training the model, we obtain the updated parameters ($\theta_{\text {unlearning }}$). The algorithm can be divided into the following five steps:

\begin{itemize}


    \item \textbf{Step 1: Neighbor Node Identification (Line \ref{algorithm3 step1}).} To begin, we identify the neighbor node set ($\mathcal{N}(v)$) of each unlearning node ($v$) by leveraging the adjacency matrix ($A$).


    \item \textbf{Step 2: Selecting Neighbor Nodes (Line \ref{algorithm3 step2}).} In this step, we curate a new set of neighbor nodes ($\mathcal{N}(v_u)$) by selecting nodes that meet two criteria: they belong to the same class as the unlearning nodes and they are not part of the training set. This selection process is captured by the following formula:
    \begin{equation}\label{selet neighbor}
        \begin{gathered}
            \mathcal{N}(v_u)=\{u \mid u \notin idx_\text{train} \wedge u \in \mathcal{N}(v) \wedge Y_u = Y_v\}
        \end{gathered}
    \end{equation}


    \item \textbf{Step 3: Mean Posterior Probability Calculation of the New Set of Neighbor Nodes (Line \ref{algorithm3 step3}).} This step involves computing the mean posterior probability ($\overline{Z'_{\text {v }} }$) for the new set of neighbor nodes ($\mathcal{N}(v_u)$) obtained in Step 2. The following equation represents the calculation process:

    \begin{equation}\label{same mean neighbor}
        \begin{gathered}
            \overline{Z'_{\text {v }} } = \frac{1}{\left|\mathcal{N}(v_u)\right|} \sum_{t}^{\left|\mathcal{N}(v_u))\right|}  Z_t, \forall t \in \mathcal{N}(v_u)
        \end{gathered}
    \end{equation}


    \item \textbf{Step 4: Modification of Unlearning Node Labels (Line \ref{algorithm3 step41} - Line \ref{algorithm3 step42}).} In this step, we proceed to update the labels of the unlearning nodes ($Y^{\prime}_v$). There are two scenarios for updating the node labels in this unlearning method. If the new set obtained in Step 3 is not empty, we update the labels of the unlearning nodes using the mean posterior probability ($\overline{Z'_{\text {v }} }$) calculated from the new set of neighbor nodes. However, if the new set obtained in Step 3 is empty (meaning there are no nodes in the neighborhood that belong to the same class as the unlearning nodes and are not part of the training set), we utilize the mean posterior probability ($\overline{Z'_{\text {c }} }$) derived from the testing set of the same class, as presented in Algorithm \ref{algorithm mean}, to update the labels of these nodes.
    \begin{equation}\label{modify same mean neighbor}
        \begin{gathered}
            Y^{\prime}_v= \begin{cases}\overline{Z'_{\text {v }} } & \text { if } {\left|\mathcal{N}(v_u)\right|} > 0 ; \\ \overline{Z'_{\text {c }} }, Y_v = c & \text { else}\end{cases}
        \end{gathered}
    \end{equation}


    \item \textbf{Step 5: Model Training (Line \ref{algorithm3 step51} - Line \ref{algorithm3 step52}).} Once we have updated the labels of the unlearning nodes in Step 4, we proceed to update the model parameters through training. This step involves iteratively updating the model parameters to minimize the loss function and improve the model's performance.

\end{itemize}

\section{Experiments}
\subsection{Experimental Setup}
\subsubsection{Datasets}
To evaluate the efficacy of node unlearning, we employ three commonly used datasets: Cora, Citeseer, and Pubmed \cite{DBLP:conf/iclr/KipfW17}. The dataset statistics are summarized in Table \ref{Datasets}.

\begin{table}[h]
  \centering
  \caption{Dataset statistics.}
        \scalebox{0.86}{
    \begin{tabular}{c|cccccc}
    \toprule
          & \textbf{Train/Val} & \textbf{Test} & \textbf{Nodes} & \textbf{Edges} & \textbf{Feats} & \textbf{Classes} \\
    \midrule
    \textbf{Cora} & 140/300 & 1,000 & 2,708 & 10,556 & 1,433 & 7 \\
    \textbf{Citeseer} & 120/500 & 1,000 & 3,327 & 9,228 & 3,703 & 6 \\
    \textbf{Pubmed} & 60/500 & 1,000 & 19,717 & 88,651 & 500   & 3 \\
    \bottomrule
    \end{tabular}%
    }
  \label{Datasets}%
\end{table}%
\subsubsection{Metrics}
In this paper, we evaluate the effectiveness of the node unlearning method from three perspectives: Model Utility, Unlearning Utility, Unlearning Utility.
\begin{itemize}

    
    \item \textbf{Model Utility.} The effectiveness of the unlearning model relies on its capacity to sustain accurate predictions after node removal. To evaluate the model's utility, we employ the widely-used accuracy metric (\textbf{Accuracy}), which assesses the multi-class classification prediction capability of GNN models.

    
    \item \textbf{Unlearning Utility.} The primary objective of unlearning techniques is to safeguard node privacy. To achieve privacy protection, it is crucial to consider the extent to which forgotten nodes are effectively removed when comparing different unlearning methods. In this study, we evaluate the success of unlearning methods in removing nodes from the training set by conducting membership inference attacks (\textbf{MIA}) \cite{DBLP:conf/sp/ShokriSSS17}. The membership inference attack is a method employed to determine whether data belongs to the training or testing set.



    \item \textbf{Unlearning Efficiency.} In addition to measuring Model Utility and Unlearning Utility, Unlearning Efficiency is also a crucial metric to consider. Unlearning methods that demonstrate high efficiency outperform the Retrain approach by significantly reducing the computational time required. To assess the efficiency of unlearning, we employ two key metrics: the number of epochs (\textbf{Epoch}) and the running time (\textbf{Time}). These metrics provide valuable insights into the computational efficiency and convergence speed of the unlearning process, enabling a comprehensive evaluation of the method's performance in terms of computational effectiveness.
    
\end{itemize}


\subsubsection{Baselines}

In the node unlearning experiments, to illustrate the effectiveness of our proposed method, we compare our method with the following model approaches.
\begin{itemize}

    
    \item \textbf{Retrain.} This is the most straightforward approach to graph unlearning. It involves removing the unlearning nodes and retraining the model from scratch using the remaining nodes. This method ensures that the influence of unlearning nodes is completely removed from the model.

    \item \textbf{Naive.} This approach involves fine-tuning the model for unlearning in a more aggressive manner. It entails updating the model by applying the inverse gradient during the node unlearning process. This method aims to quickly adapt the model parameters to remove the influence of the unlearning nodes.

    \item \textbf{GraphEraser \cite{DBLP:conf/ccs/Chen000H022}.} This study presents a novel machine unlearning method specifically designed for graph data, which extends and adapts the SISA method to graph-based applications. The proposed approach entails dividing the entire training data into multiple shards and training a sub-model for each shard. When an unlearning request is received, only the sub-models associated with the shards containing the unlearned data need to be retrained. The final prediction is obtained by aggregating the predictions from all the sub-models.

    \item \textbf{GIF \cite{DBLP:conf/www/WuYQS0023}.} This study explores the utilization of influence functions customized for graph unlearning. The core concept involves extending the traditional influence function by introducing an additional loss term that captures the influence on neighboring nodes. By deriving closed-form solutions for parameter adjustments, the researchers gain a deeper understanding of the underlying mechanism of unlearning in graph-based models.

\end{itemize}

\subsubsection{Settings}

To evaluate the effectiveness of our proposed node unlearning approach, we conducted experiments on Graph Convolutional Network (GCN)  \cite{DBLP:conf/iclr/KipfW17}, Simplified Graph Convolutional Network (SGC) \cite{DBLP:conf/icml/WuSZFYW19} and ARMA Graph Neural Network (ARMA) \cite{DBLP:journals/pami/BianchiGLA22}. All models consisted of two-layer networks. The activation function used for the hidden layer was ReLU, while the output layer utilized the Softmax activation function. The optimization algorithm employed was Adam, with a learning rate set to $0.001$. The models were trained for a total of $1600$ epochs.

As presented in Table \ref{Datasets}, the commonly used node classification datasets do not provide a sufficient number of original training nodes to facilitate the study of node unlearning. To overcome this limitation, we augmented the training dataset by incorporating a portion of the validation set for our experiments. In addition to the data merging step, we conducted various node unlearning experiments on the combined training set. Specifically, we selectively removed different percentages of nodes from the training set, including $20\%$, $40\%$, $60\%$, and $80\%$, to evaluate the impact of node unlearning on the model's performance.


\subsection{Performance Comparison}

\subsubsection{Model's Utility}\label{Comparison1}
To begin our evaluation, we assess the utility of the three proposed approaches after node unlearning. We conduct experiments using the GCN model on the three datasets and compare the results with baseline methods. The outcomes of these experiments are summarized in Table \ref{Accuracy results}, providing a comprehensive overview of the model's performance on different datasets.


The Naive technique utilizes the negative gradient of the unlearning node to directly optimize the model for node unlearning. However, as indicated in Table \ref{Accuracy results}, this approach significantly compromises the utility of the model. The models generated through the Naive method exhibit poor performance in predicting the classification of nodes across all three datasets. The accuracy rates in Cora, Citeseer, and Pubmed are comparable to random guessing probabilities.



Our proposed method effectively preserves the model's utility, as evidenced by the results presented in the table. Notably, the unlearning technique that incorporates graph topology outperforms the Class-based Label Replacement (CLR) method, which solely relies on the mean posterior of the testing set. Additionally, the Topology-guided Neighbor Mean Posterior Probability (TNMPP) method achieves the highest accuracy rate, reaching up to $80.75\%$ node classification accuracy in the best-case scenario. It proves to be more practical than the Class-consistent Neighbor Node Filtering (CNNF) method. The superiority of the TNMPP technique can be attributed to the inclusion of neighboring nodes belonging to the training set, thereby enhancing the model's utility while slightly compromising the effectiveness of unlearning. In the following section, we will delve further into the experimental analysis of this phase.

\begin{table}[tbp]
  \centering
  \caption{The experimental results for each model with different numbers of unlearning nodes.}
        \scalebox{0.72}{
    \begin{tabular}{cccccc}
    \toprule
    \textbf{Dataset} & \textbf{Method} & \textbf{20\%} & \textbf{40\%} & \textbf{60\%} & \textbf{80\%} \\
    \midrule
    \multirow{5}[2]{*}{\textbf{Cora}} & \textbf{Retrain} & 79.25±0.73 & 74.45±0.79 & 74.23±0.85 & 73.48±0.91 \\
          & \textbf{Naive} & 22.93±10.39 & 19.03±8.59 & 22.80±10.52 & 19.15±8.50 \\
          & \textbf{CLR} & 79.63±0.39 & 78.25±0.71 & 76.70±0.70 & 76.23±1.17 \\
          & \textbf{TNMPP} & 80.75±0.47 & 78.90±0.47 & 78.65±1.64 & 78.10±0.36 \\
          & \textbf{CNNF} & 79.88±0.34 & 78.70±0.71 & 77.35±0.78 & 76.48±1.12 \\
    \midrule
    \multirow{5}[2]{*}{\textbf{Citeseer}} & \textbf{Retrain} & 68.06±0.70 & 64.16±1.29 & 62.93±1.01 & 61.86±1.02 \\
          & \textbf{Naive} & 9.78±4.65 & 13.94±5.70 & 19.35±2.50 & 18.86±2.43 \\
          & \textbf{CLR} & 68.12±0.29 & 66.94±0.21 & 65.70±0.85 & 64.76±0.98 \\
          & \textbf{TNMPP} & 69.96±0.59 & 68.84±0.50 & 68.38±0.26 & 67.50±0.59 \\
          & \textbf{CNNF} & 68.44±0.26 & 67.52±0.23 & 66.40±1.11 & 65.22±0.89 \\
    \midrule
    \multirow{5}[2]{*}{\textbf{Pubmed}} & \textbf{Retrain} & 75.73±0.71 & 75.22±0.55 & 70.62±0.51 & 69.90±1.15 \\
          & \textbf{Naive} & 37.42±9.51 & 22.66±10.42 & 36.64±10.42 & 33.53±12.03 \\
          & \textbf{CLR} & 76.17±0.08 & 75.74±0.56 & 74.58±0.83 & 73.92±0.54 \\
          & \textbf{TNMPP} & 78.07±0.50 & 77.56±0.48 & 77.48±0.17 & 76.90±0.83 \\
          & \textbf{CNNF} & 76.58±0.21 & 76.06±0.47 & 74.90±0.93 & 74.63±0.50 \\
    \bottomrule
    \end{tabular}%
    }
  \label{Accuracy results}%
\end{table}%


We also conducted unlearning experiments with varying numbers of nodes, specifically $20\%$, $40\%$, $60\%$, and $80\%$ of the total number of training nodes. The results presented in the three tables demonstrate that our proposed method maintains good model utility across different degrees of unlearning, achieving performance levels comparable to the Retrain method.

\begin{figure*}[t!]
    \subfigure[Cora]{
    \begin{minipage}[t]{0.31\textwidth}
           \centering
           \includegraphics[width=\textwidth]{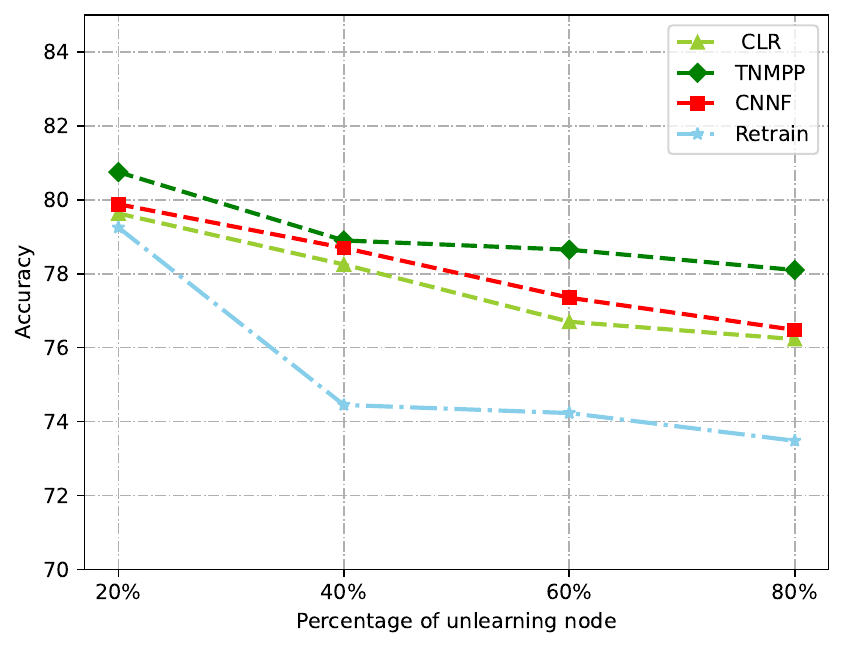}
    \end{minipage}
    }
    \subfigure[Citeseer]{
    \begin{minipage}[t]{0.31\textwidth}
            \centering
            \includegraphics[width=\textwidth]{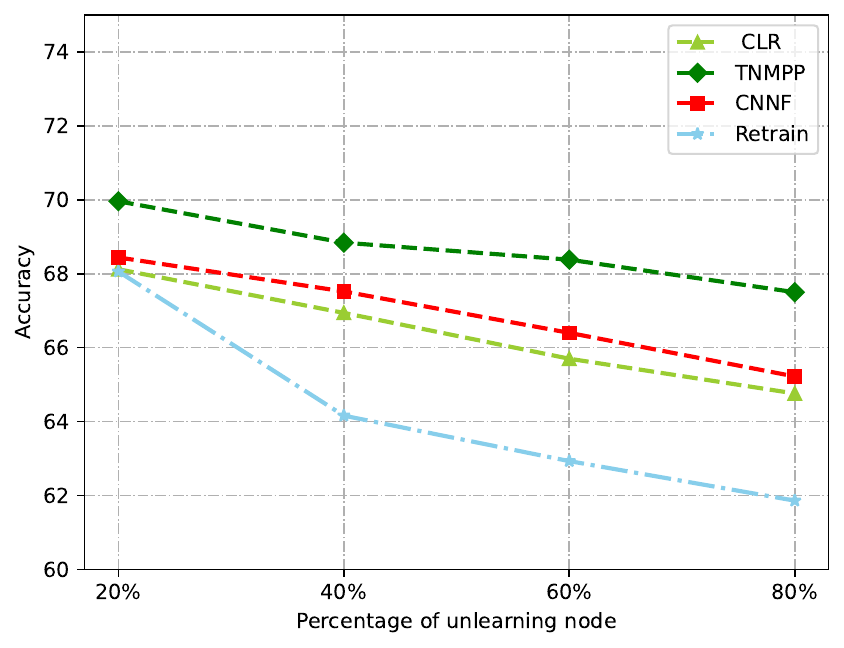}
    \end{minipage}
    }
    \subfigure[Pubmed]{
    \begin{minipage}[t]{0.31\textwidth}
            \centering
            \includegraphics[width=\textwidth]{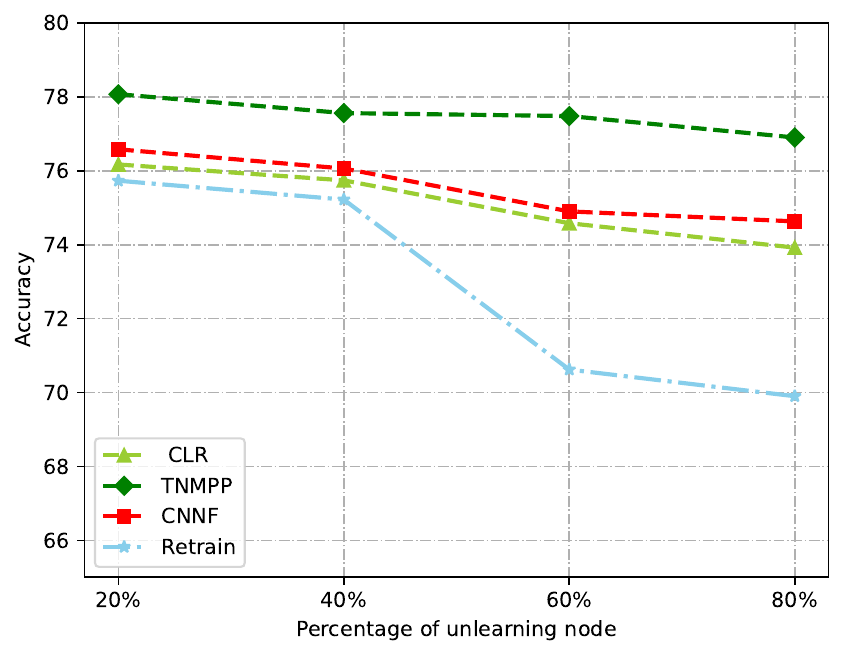}
    \end{minipage}
    }
\caption{Experimental comparison curves for each model using different numbers of nodes for unlearning.}
\label{com_res}
\end{figure*}

Furthermore, in machine learning, it is well-known that an increase in the number of training nodes generally leads to higher model accuracy. Fig. \ref{com_res} illustrates that as the number of unlearning nodes increases, the model's utility gradually diminishes, which aligns with fundamental machine learning principles.

\begin{figure*}[t!]
    \subfigure[Cora]{
    \begin{minipage}[t]{0.31\textwidth}
           \centering
           \includegraphics[width=\textwidth]{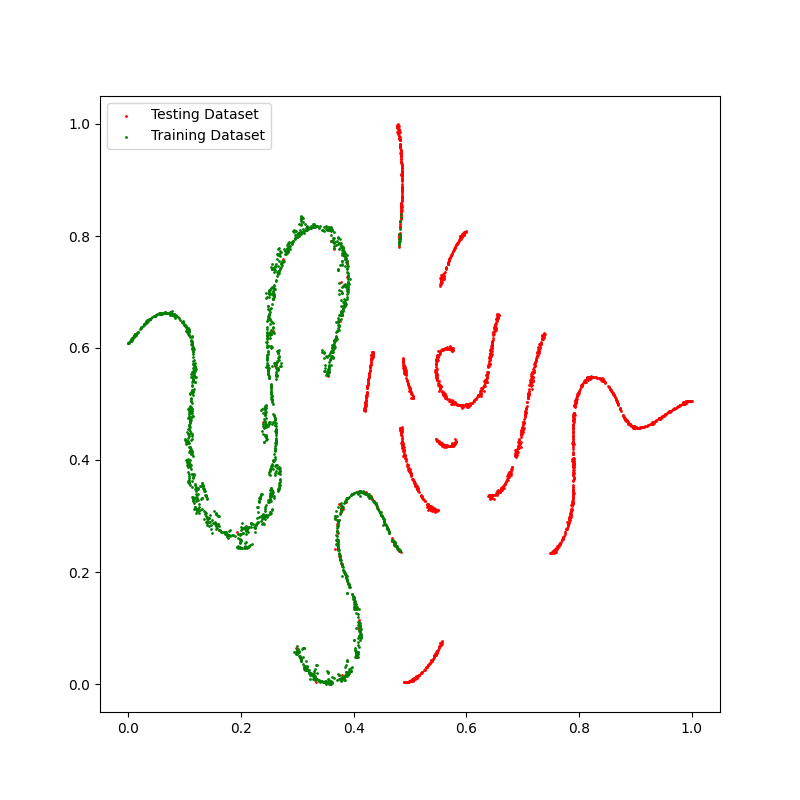}
    \end{minipage}
    }
    \subfigure[Citeseer]{
    \begin{minipage}[t]{0.31\textwidth}
            \centering
            \includegraphics[width=\textwidth]{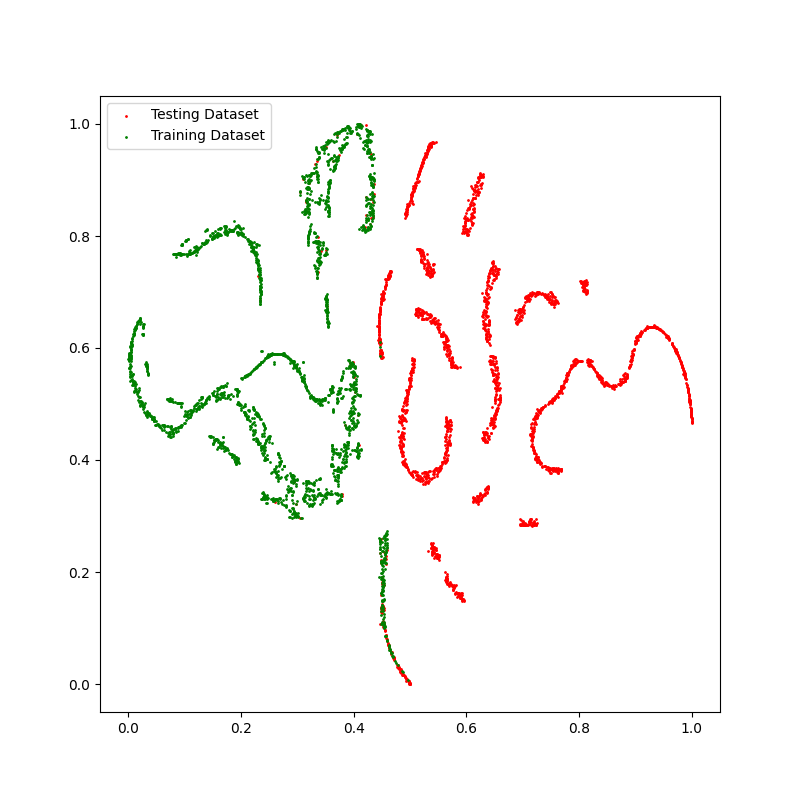}
    \end{minipage}
    }
    \subfigure[Pubmed]{
    \begin{minipage}[t]{0.31\textwidth}
            \centering
            \includegraphics[width=\textwidth]{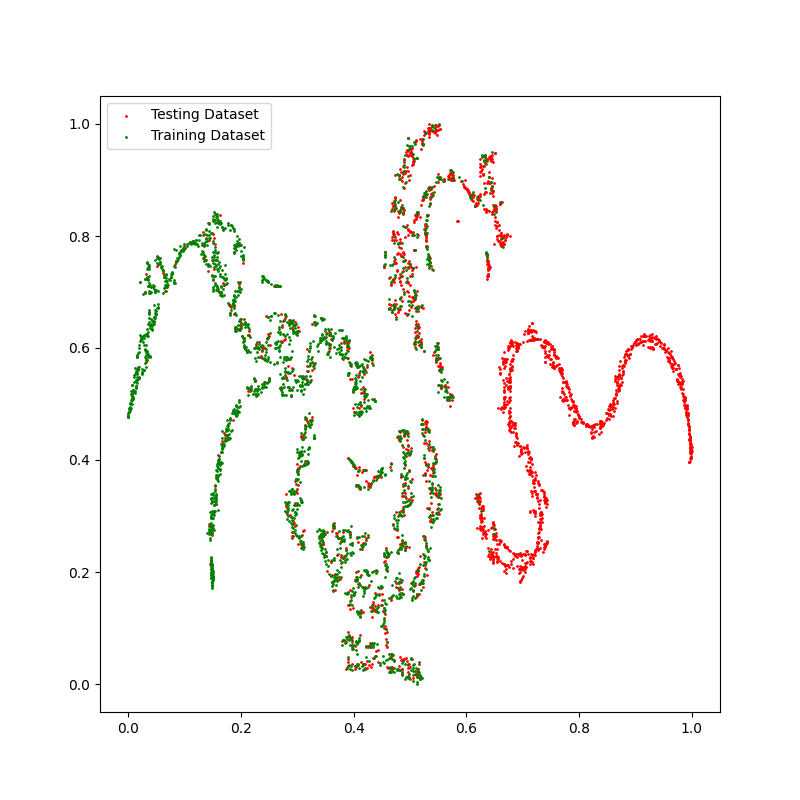}
    \end{minipage}
    }
\caption{T-SNE diagram of member inference attack.}
\label{MIA Base}
\end{figure*}
 
\subsubsection{Unlearning Utility}

Unlearning plays a crucial role in mitigating the impact of specific data or features on the model. By unlearning, we effectively prevent attackers from exploiting the model to access sensitive data, thereby protecting user privacy. In order to gain insights into the effectiveness of node unlearning, we visualize the posterior probability distributions of the training and testing sets, as depicted in Fig. \ref{MIA Base}. Notably, there is a noticeable disparity between the posterior probabilities of the training set and the testing set. Through unlearning, we can remove the relevant data from the training set while ensuring that the predictive performance of the unlearning data is consistent with the testing data. Therefore, the success of node unlearning can be assessed by whether the node successfully shifts from the distribution of the training set to the distribution of the testing set. To further evaluate the efficacy of unlearning, we employ membership inference attacks (MIA) to determine whether a given node belongs to the training set or the testing set.


Ideally, if the influence of the unlearning node is successfully removed, the membership inference attack should predict that the unlearning node does not belong to the training dataset. To evaluate the unlearning effectiveness of the proposed node unlearning methods, we compare their performance in Table \ref{MIA results}. The experimental results demonstrate the efficacy of all three node unlearning methods. Even when removing varying percentages of training nodes $20\%$, $40\%$, $60\%$, $80\%$, the methods proposed in this research consistently achieve MIA accuracy levels above $85\%$ on the Cora and Citeseer datasets in the GCN model. It is worth noting that the simple MIA model utilized in this study performs relatively less effectively on the Pubmed dataset compared to the other two datasets. However, the MIA accuracy of our proposed method on the Pubmed dataset still exceeds $70\%$ for each percentage of unlearning. In summary, all of the methods presented in this paper successfully accomplish node removal, as substantiated by the experimental results.



\begin{figure*}[t!]
    \subfigure[Cora]{
    \begin{minipage}[t]{0.31\textwidth}
           \centering
           \includegraphics[width=\textwidth]{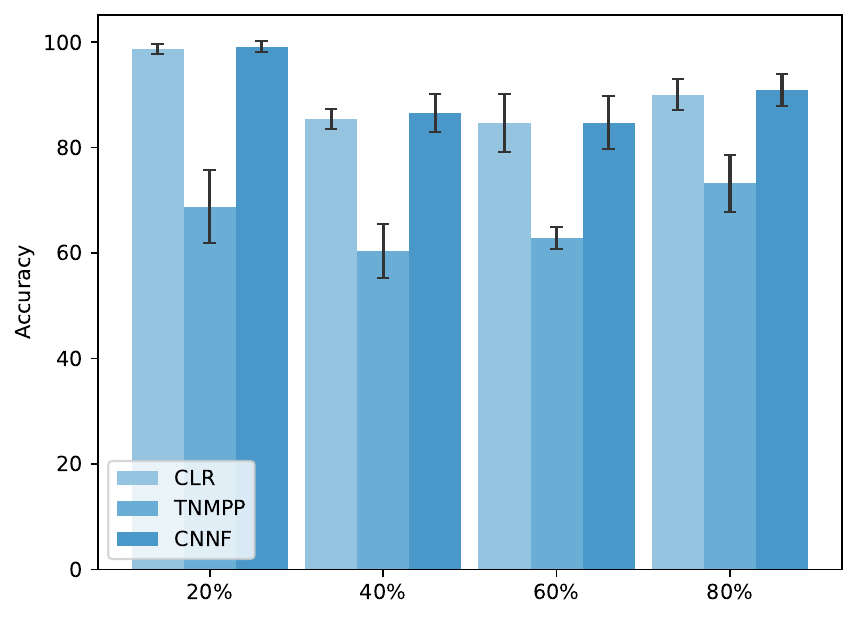}
    \end{minipage}
    }
    \subfigure[Citeseer]{
    \begin{minipage}[t]{0.31\textwidth}
            \centering
            \includegraphics[width=\textwidth]{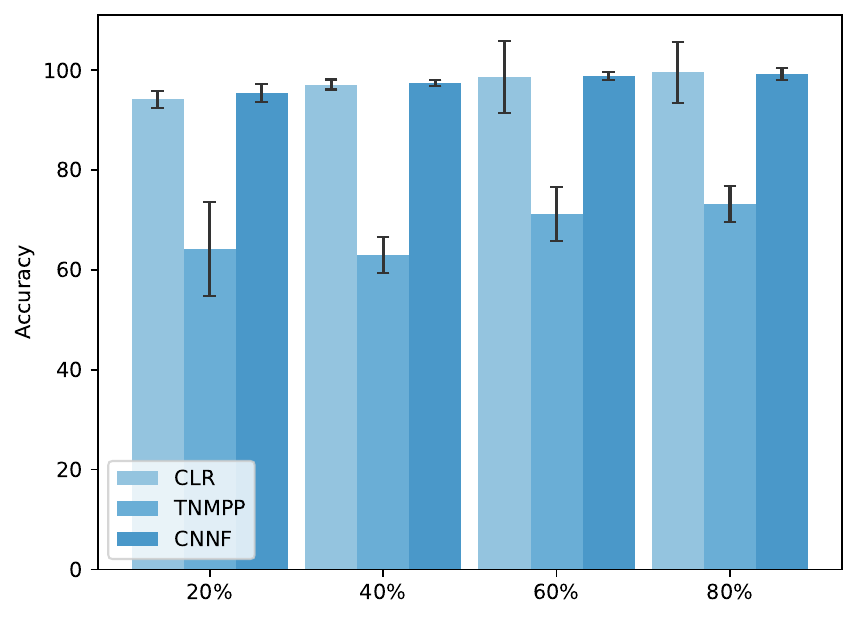}
    \end{minipage}
    }
    \subfigure[Pubmed]{
    \begin{minipage}[t]{0.31\textwidth}
            \centering
            \includegraphics[width=\textwidth]{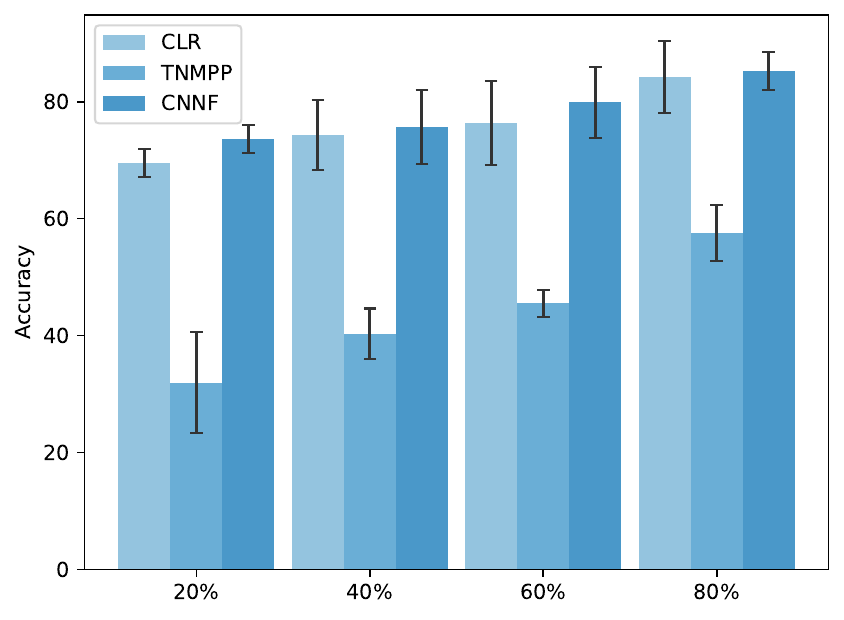}
    \end{minipage}
    }
\caption{Comparison of membership inference attack accuracy for unlearning nodes using our proposed method.}
\label{com_unlearning}
\end{figure*}

The effectiveness of unlearning varies among the three methods proposed in this research. On all three datasets, the unlearning efficiency of the TNMPP method is lower than that of the other two methods, regardless of the proportion of unlearning nodes, as depicted in Table \ref{MIA results}. The TNMPP method achieves better model utility due to its utilization of some training nodes in the computation of the mean posterior probability, as mentioned in the previous section \ref{Comparison1}. However, the presence of training nodes also leads to the poorest unlearning utility, as the training nodes become mixed during the replacement of the target class of the unlearning nodes, hindering complete node forgetting. The MIA performance of the unlearning nodes alone is presented in Table \ref{MIA results}, revealing that the MIA accuracy of the unlearning nodes in the TNMPP method is at least $15\%$ lower compared to all other methods. To illustrate the superiority of CNNF over TNMPP in terms of unlearning utility, we visualize the comparison of the MIA accuracy of the proposed method on the unlearning nodes, as shown in Fig. \ref{com_unlearning}. As depicted in the figure, the unlearning utility of CNNF significantly surpasses TNMPP.

\begin{table*}[t!]
  \centering
  \caption{The accuracy of MIA for each model with different numbers of unlearning nodes. The term \emph{All Node} refers to the MIA results of all nodes, while \emph{Unlearning Node} pertains to the MIA results of the unlearning nodes.}
        \scalebox{0.86}{
    \begin{tabular}{ccccccccccc}
    \toprule
    \multirow{2}[4]{*}{\textbf{Model}} & \multirow{2}[4]{*}{\textbf{Dataset}} & \multirow{2}[4]{*}{\textbf{Method}} & \multicolumn{4}{c}{\textbf{All Node}} & \multicolumn{4}{c}{\textbf{Unlearning Node}} \\
\cmidrule{4-11}          &       &       & \textbf{20\%} & \textbf{40\%} & \textbf{60\%} & \textbf{80\%} & \textbf{20\%} & \textbf{40\%} & \textbf{60\%} & \textbf{80\%} \\
    \midrule
    \multirow{9}[6]{*}{\textbf{GCN}} & \multirow{3}[2]{*}{\textbf{Cora}} & \textbf{CLR} & 87.95±0.90 & 88.54±0.52 & 90.16±1.18 & 93.72±1.00 & 98.66±0.90 & 85.42±1.86 & 84.72±5.53 & 90.03±2.97 \\
          &       & \textbf{TNMPP} & 85.57±1.40 & 85.10±1.01 & 86.49±1.02 & 88.51±0.95 & 68.75±6.92 & 60.41±5.16 & 62.80±2.11 & 73.21±5.36 \\
          &       & \textbf{CNNF} & 87.93±1.16 & 88.38±0.76 & 90.26±1.07 & 93.88±0.76 & 99.11±1.03 & 86.61±3.57 & 84.72±4.99 & 90.92±2.97 \\
\cmidrule{2-11}          & \multirow{3}[2]{*}{\textbf{Citeseer}} & \textbf{CLR} & 91.18±1.50 & 92.30±0.23 & 93.90±0.39 & 97.34±0.37 & 94.17±1.74 & 97.14±1.00 & 98.61±0.70 & 99.61±0.26 \\
          &       & \textbf{TNMPP} & 89.07±1.48 & 87.26±0.51 & 88.76±2.96 & 89.10±2.18 & 64.17±9.36 & 63.02±3.66 & 71.18±5.40 & 73.18±3.65 \\
          &       & \textbf{CNNF} & 91.40±1.28 & 92.56±0.19 & 94.33±0.31 & 97.44±0.29 & 95.42±1.74 & 97.40±0.60 & 98.85±0.80 & 99.27±1.14 \\
\cmidrule{2-11}          & \multirow{3}[2]{*}{\textbf{Pubmed}} & \textbf{CLR} & 80.15±1.13 & 82.41±1.65 & 84.49±1.44 & 91.39±1.47 & 69.44±2.40 & 74.30±6.02 & 76.39±7.17 & 84.17±6.14 \\
          &       & \textbf{TNMPP} & 77.50±2.16 & 76.70±1.90 & 76.65±1.75 & 80.87±2.07 & 31.95±8.67 & 40.28±4.34 & 45.49±2.37 & 57.50±4.80 \\
          &       & \textbf{CNNF} & 80.68±0.75 & 82.86±1.59 & 84.93±1.40 & 91.53±1.23 & 73.61±2.41 & 75.69±6.36 & 79.86±6.05 & 85.21±3.24 \\
    \midrule
    \multirow{9}[6]{*}{\textbf{SGC}} & \multirow{3}[2]{*}{\textbf{Cora}} & \textbf{CLR} & 88.11±0.93 & 88.12±0.49 & 90.13±0.18 & 95.41±0.25 & 98.57±0.80 & 86.31±0.52 & 87.30±3.28 & 91.81±2.07 \\
          &       & \textbf{TNMPP} & 86.23±0.71 & 86.12±0.99 & 86.02±0.43 & 88.44±1.54 & 72.86±5.11 & 63.99±5.68 & 70.64±6.64 & 72.32±5.26 \\
          &       & \textbf{CNNF} & 88.40±0.84 & 88.73±0.53 & 90.63±0.28 & 95.55±0.34 & 98.57±0.80 & 87.20±1.86 & 88.49±3.28 & 93.15±2.46 \\
\cmidrule{2-11}          & \multirow{3}[2]{*}{\textbf{Citeseer}} & \textbf{CLR} & 89.79±0.36 & 92.23±0.26 & 93.51±0.48 & 96.37±0.35 & 94.58±1.14 & 97.57±0.60 & 97.92±0.98 & 98.61±0.60 \\
          &       & \textbf{TNMPP} & 87.64±1.06 & 87.53±1.22 & 86.31±1.58 & 87.82±5.74 & 59.17±6.69 & 68.06±5.92 & 67.36±6.61 & 64.93±3.91 \\
          &       & \textbf{CNNF} & 90.10±0.73 & 92.47±0.44 & 93.87±0.33 & 96.58±0.26 & 95.41±0.93 & 97.92±0.00 & 98.27±0.40 & 98.96±0.52 \\
\cmidrule{2-11}          & \multirow{3}[2]{*}{\textbf{Pubmed}} & \textbf{CLR} & 80.29±2.02 & 82.23±1.44 & 84.82±1.86 & 91.72±1.07 & 70.83±9.32 & 78.47±3.18 & 83.33±2.41 & 81.67±4.75 \\
          &       & \textbf{TNMPP} & 77.18±2.69 & 78.33±1.62 & 76.01±2.32 & 78.23±4.11 & 28.33±6.18 & 45.14±9.39 & 38.89±6.05 & 50.83±9.67 \\
          &       & \textbf{CNNF} & 80.61±1.89 & 83.27±1.79 & 85.09±2.30 & 92.34±0.89 & 73.33±9.59 & 81.94±1.20 & 84.72±1.39 & 84.58±3.41 \\
    \midrule
    \multirow{9}[6]{*}{\textbf{ARMA}} & \multirow{3}[2]{*}{\textbf{Cora}} & \textbf{CLR} & 79.90±0.23 & 85.68±0.55 & 89.27±0.93 & 95.19±0.72 & 97.02±2.73 & 85.42±1.36 & 88.49±3.59 & 94.82±1.36 \\
          &       & \textbf{TNMPP} & 73.49±1.01 & 74.06±1.42 & 72.99±1.61 & 73.39±3.15 & 25.00±4.73 & 30.66±4.02 & 37.90±4.05 & 42.41±4.32 \\
          &       & \textbf{CNNF} & 80.39±0.61 & 85.80±0.63 & 89.82±0.39 & 95.67±0.67 & 99.40±1.03 & 85.71±1.79 & 90.68±2.09 & 96.07±1.39 \\
\cmidrule{2-11}          & \multirow{3}[2]{*}{\textbf{Citeseer}} & \textbf{CLR} & 89.84±1.05 & 93.39±1.28 & 94.73±0.61 & 97.60±0.27 & 98.61±1.20 & 98.34±0.93 & 97.46±0.80 & 99.48±0.42 \\
          &       & \textbf{TNMPP} & 86.51±2.18 & 84.50±2.25 & 82.74±2.18 & 82.26±1.45 & 59.03±8.42 & 46.25±5.92 & 47.46±5.12 & 48.31±2.74 \\
          &       & \textbf{CNNF} & 90.06±0.93 & 93.53±1.24 & 95.02±0.62 & 97.84±0.22 & 99.31±1.20 & 98.75±0.47 & 97.92±0.70 & 99.61±0.26 \\
\cmidrule{2-11}          & \multirow{3}[2]{*}{\textbf{Pubmed}} & \textbf{CLR} & 75.95±0.83 & 79.76±0.34 & 83.78±1.44 & 91.10±0.90 & 76.39±2.41 & 79.86±1.20 & 81.94±1.39 & 88.19±1.59 \\
          &       & \textbf{TNMPP} & 72.56±0.99 & 72.83±1.33 & 70.21±1.26 & 69.88±4.16 & 48.61±4.82 & 42.36±9.84 & 43.52±3.49 & 53.13±2.76 \\
          &       & \textbf{CNNF} & 76.49±0.21 & 80.68±0.36 & 84.79±1.94 & 91.46±0.72 & 81.94±2.40 & 82.64±1.20 & 83.33±2.78 & 88.54±1.80 \\
    \bottomrule
    \end{tabular}%
    }
  \label{MIA results}%
\end{table*}%

\begin{figure*}[t!]
    \subfigure[Cora]{
    \begin{minipage}[t]{0.31\textwidth}
           \centering
           \includegraphics[width=\textwidth]{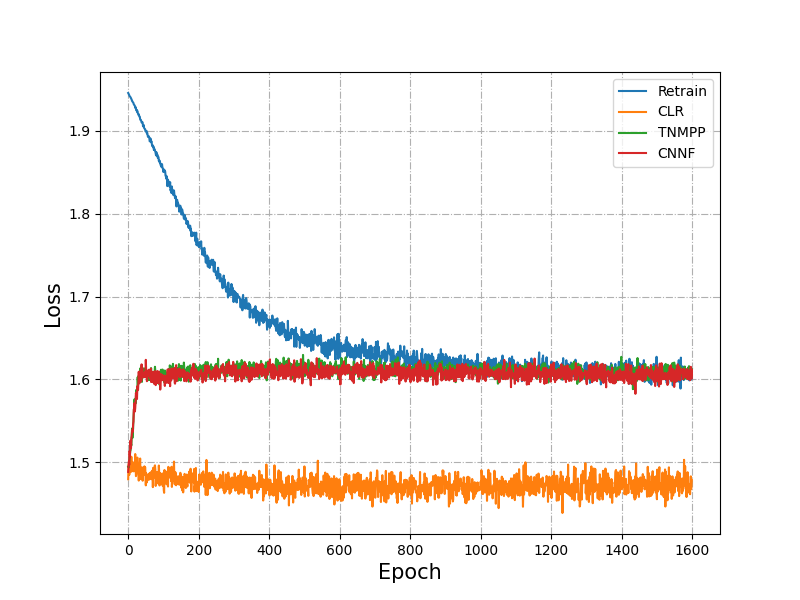}
    \end{minipage}
    }
    \subfigure[Citeseer]{
    \begin{minipage}[t]{0.31\textwidth}
            \centering
            \includegraphics[width=\textwidth]{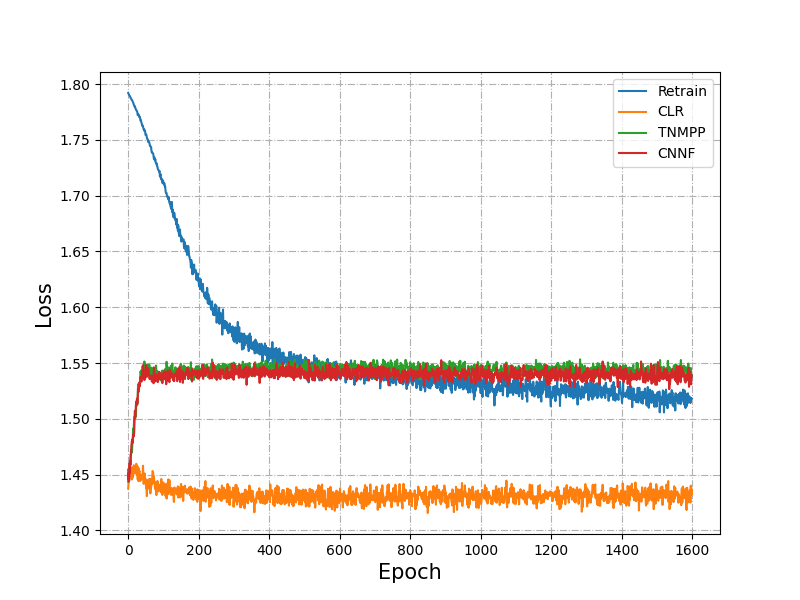}
    \end{minipage}
    }
    \subfigure[Pubmed]{
    \begin{minipage}[t]{0.31\textwidth}
            \centering
            \includegraphics[width=\textwidth]{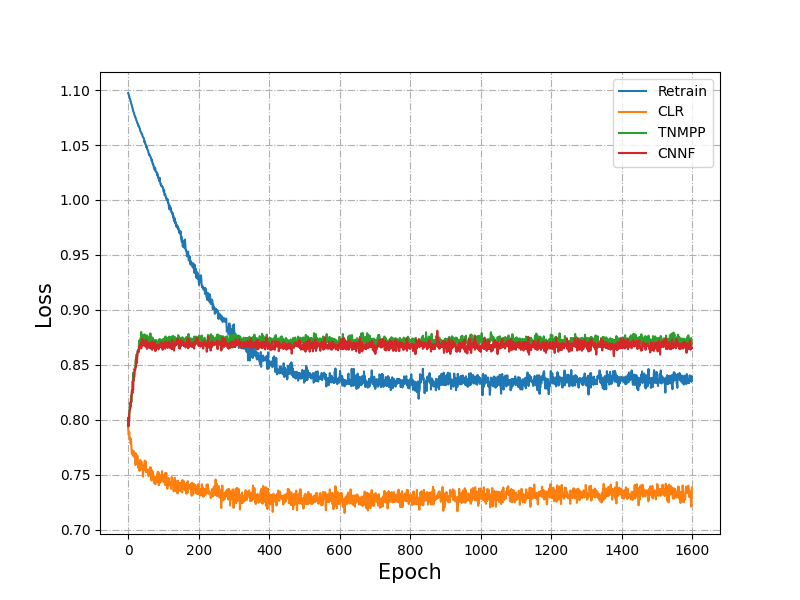}
    \end{minipage}
    }
\caption{Each model's training process in the different dataset. Where the x-axis indicates epoch and the y-axis indicates losses.}
\label{training process}
\end{figure*}

We further investigated the relevance of topology to the node forgetting performance in graph unlearning. By comparing the CLR method with the CNNF method in Table \ref{MIA results}, it is evident that the CNNF method consistently surpasses the CLR method in terms of MIA accuracy across all experiments. This suggests that the CNNF method exhibits superior node unlearning compared to the CLR method. Consequently, incorporating topological structure features not only enhances the model's utility but also improves the effectiveness of unlearning.


\begin{figure*}[t!]
    \subfigure[Cora]{
    \begin{minipage}[t]{0.31\textwidth}
           \centering
           \includegraphics[width=\textwidth]{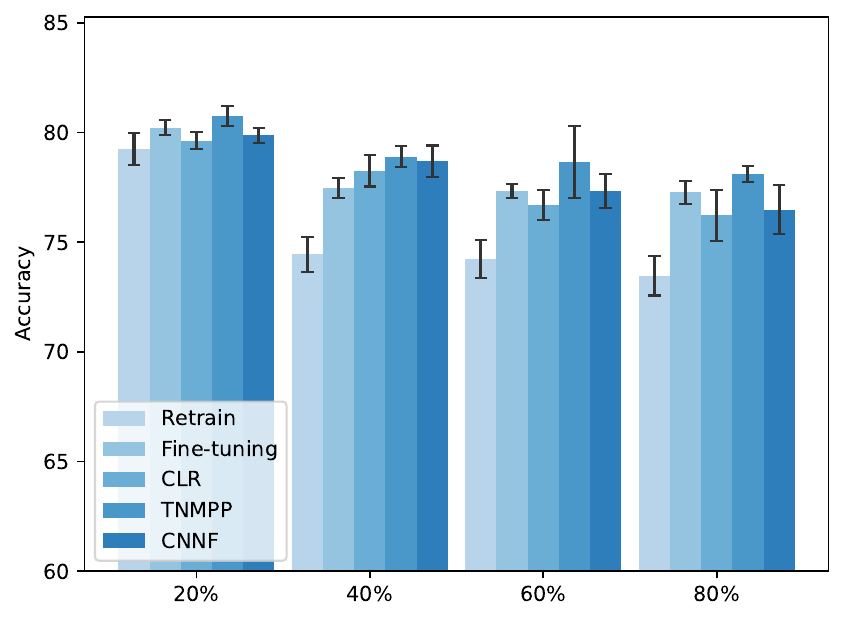}
    \end{minipage}
    }
    \subfigure[Citeseer]{
    \begin{minipage}[t]{0.31\textwidth}
            \centering
            \includegraphics[width=\textwidth]{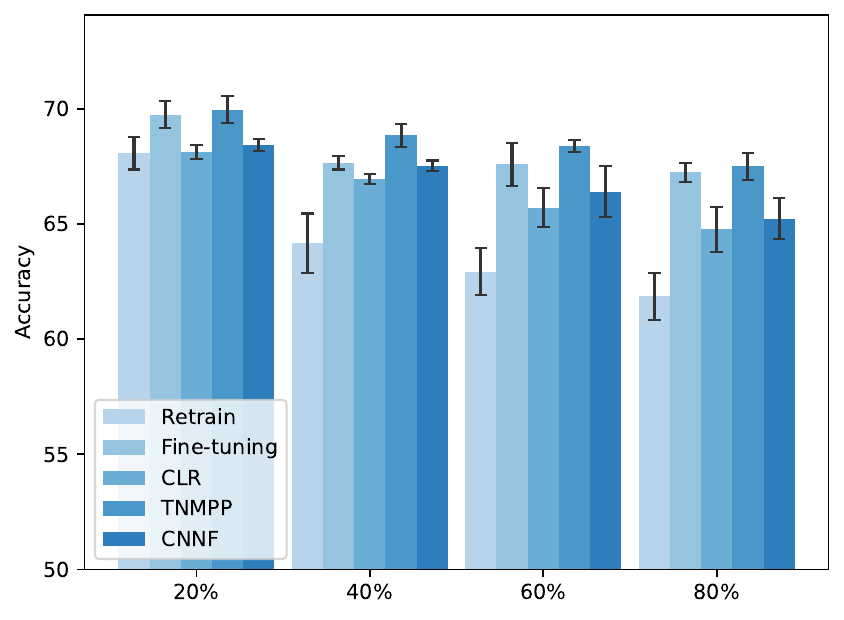}
    \end{minipage}
    }
    \subfigure[Pubmed]{
    \begin{minipage}[t]{0.31\textwidth}
            \centering
            \includegraphics[width=\textwidth]{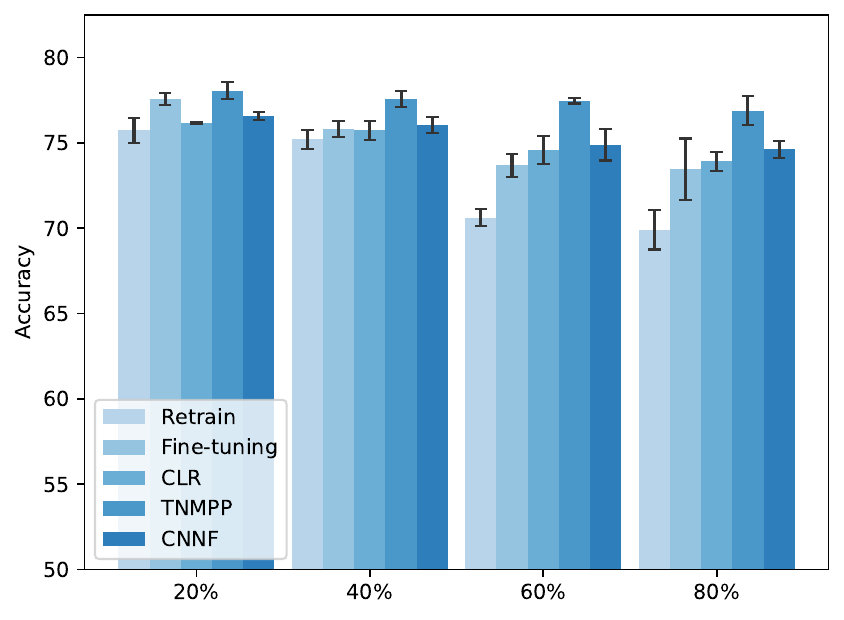}
    \end{minipage}
    }
\caption{Comparison of model accuracy for different methods.}
\label{com_finetuning}
\end{figure*}
\subsubsection{Unlearning's Efficiency}

By selectively removing specific training data that needs to be forgotten, unlearning methods contribute significantly to privacy preservation objectives. While the Retrain model also serves the purpose of privacy protection, it can be computationally expensive and time-consuming, especially when dealing with a large number of unlearning nodes. Therefore, to highlight the advantages of unlearning, it is crucial for the unlearning process to outperform the Retrain approach in terms of training efficiency. This efficiency ensures that unlearning achieves the desired privacy outcomes in a more time-efficient manner compared to the process of retraining the entire model.

We present the training loss plots for each model in Fig. \ref{training process}. The plots clearly demonstrate that our strategy achieves model convergence in fewer than $70$ epochs, while the Retrain strategy requires approximately $500$ epochs to achieve comparable model stability. These experimental results provide compelling evidence that our approach is significantly more efficient in training the model across all datasets compared to the Retrain method. Consequently, in this study, unlearning proves to be over $7$ times more effective than Retrain in terms of training efficiency.

In addition, we provide the running time comparison between our method and the Retrain approach in Table \ref{com_pre}. The results indicate that our method consistently achieves a running time of less than $1$ second, regardless of the dataset and data settings, while the Retrain method requires approximately $4$ to $6$ seconds to complete. These findings clearly demonstrate the superior efficiency of our method compared to Retrain.




\subsubsection{Fine-tuning and Retrain Models}

It is noteworthy that our proposed model consistently outperforms the Retrain method across various scenarios. In the best-case scenario, our method surpasses Retrain by nearly $5\%$. The inferior performance of the Retrain model can be attributed to its approach of building a new model from scratch. This method is susceptible to underfitting, particularly when the dataset is limited, resulting in suboptimal model performance.

In contrast, our proposed fine-tuning model leverages a pre-trained model and requires minimal additional data for adjustment, leading to significant improvements. By employing fine-tuning in our suggested approaches, our strategy efficiently adapts the pre-trained features to new data with minimal training requirements. Consequently, the Fine-tuning model outperforms the Retrain model, especially in situations where the training dataset is too small to effectively train the model.

We conducted validation experiments to understand why our proposed strategy is superior to the Retrain model. We accomplished this by fine-tuning the pre-trained models directly in addition to the unlearning node, as illustrated in Fig. \ref{com_finetuning}. The results show that fine-tuning outperforms the Retrain method and achieves a similar level of accuracy to our proposed method. Notably, the model obtained through pre-training surpasses the Retrain model while requiring less training data. This experiment confirms the fundamental reason why our method must outperform Retrain.

\begin{table}[t!]
  \centering
  \caption{The experimental results for each model with different numbers of unlearning nodes in SGC and ARMA model.}
          \scalebox{0.63}{
    \begin{tabular}{rrrrrrr}
    \toprule
    \multicolumn{1}{c}{\textbf{Model}} & \multicolumn{1}{c}{\textbf{Dataset}} & \multicolumn{1}{c}{\textbf{Method}} & \multicolumn{1}{c}{\textbf{20\%}} & \multicolumn{1}{c}{\textbf{40\%}} & \multicolumn{1}{c}{\textbf{60\%}} & \multicolumn{1}{c}{\textbf{80\%}} \\
    \midrule
    \multicolumn{1}{c}{\multirow{15}[6]{*}{\textbf{SGC}}} & \multicolumn{1}{c}{\multirow{5}[2]{*}{\textbf{Cora}}} & \multicolumn{1}{c}{\textbf{Retrain}} & \multicolumn{1}{c}{78.90±0.67} & \multicolumn{1}{c}{76.02±0.62} & \multicolumn{1}{c}{74.51±0.98} & \multicolumn{1}{c}{70.12±0.75} \\
          &       & \multicolumn{1}{c}{\textbf{Naive}} & \multicolumn{1}{c}{27.65±8.50} & \multicolumn{1}{c}{14.42±0.74} & \multicolumn{1}{c}{16.49±10.07} & \multicolumn{1}{c}{16.26±8.98} \\
          &       & \multicolumn{1}{c}{\textbf{CLR}} & \multicolumn{1}{c}{79.80±0.45} & \multicolumn{1}{c}{77.92±0.73} & \multicolumn{1}{c}{76.63±0.98} & \multicolumn{1}{c}{75.80±0.55} \\
          &       & \multicolumn{1}{c}{\textbf{TNMPP}} & \multicolumn{1}{c}{80.65±0.51} & \multicolumn{1}{c}{78.62±0.50} & \multicolumn{1}{c}{77.83±0.59} & \multicolumn{1}{c}{78.32±0.63} \\
          &       & \multicolumn{1}{c}{\textbf{CNNF}} & \multicolumn{1}{c}{80.08±0.49} & \multicolumn{1}{c}{78.30±0.77} & \multicolumn{1}{c}{77.11±1.09} & \multicolumn{1}{c}{76.26±0.47} \\
\cmidrule{2-7}          & \multicolumn{1}{c}{\multirow{5}[2]{*}{\textbf{Citeseer}}} & \multicolumn{1}{c}{\textbf{Retrain}} & \multicolumn{1}{c}{68.16±0.74} & \multicolumn{1}{c}{64.12±0.70} & \multicolumn{1}{c}{63.43±0.36} & \multicolumn{1}{c}{61.96±0.66} \\
          &       & \multicolumn{1}{c}{\textbf{Naive}} & \multicolumn{1}{c}{11.88±5.72} & \multicolumn{1}{c}{17.62±0.66} & \multicolumn{1}{c}{20.63±2.86} & \multicolumn{1}{c}{17.48±0.99} \\
          &       & \multicolumn{1}{c}{\textbf{CLR}} & \multicolumn{1}{c}{68.44±0.21} & \multicolumn{1}{c}{67.00±0.86} & \multicolumn{1}{c}{65.53±0.49} & \multicolumn{1}{c}{65.06±1.31} \\
          &       & \multicolumn{1}{c}{\textbf{TNMPP}} & \multicolumn{1}{c}{69.72±0.61} & \multicolumn{1}{c}{68.42±1.00} & \multicolumn{1}{c}{68.25±0.44} & \multicolumn{1}{c}{67.08±0.64} \\
          &       & \multicolumn{1}{c}{\textbf{CNNF}} & \multicolumn{1}{c}{68.78±0.30} & \multicolumn{1}{c}{67.50±1.06} & \multicolumn{1}{c}{66.00±0.38} & \multicolumn{1}{c}{65.40±1.25} \\
\cmidrule{2-7}          & \multicolumn{1}{c}{\multirow{5}[2]{*}{\textbf{Pubmed}}} & \multicolumn{1}{c}{\textbf{Retrain}} & \multicolumn{1}{c}{76.10±0.92} & \multicolumn{1}{c}{73.93±1.06} & \multicolumn{1}{c}{70.90±0.33} & \multicolumn{1}{c}{69.58±2.35} \\
          &       & \multicolumn{1}{c}{\textbf{Naive}} & \multicolumn{1}{c}{34.64±11.37} & \multicolumn{1}{c}{23.83±11.65} & \multicolumn{1}{c}{37.32±9.47} & \multicolumn{1}{c}{35.48±11.65} \\
          &       & \multicolumn{1}{c}{\textbf{CLR}} & \multicolumn{1}{c}{76.73±0.30} & \multicolumn{1}{c}{75.48±0.67} & \multicolumn{1}{c}{74.68±1.17} & \multicolumn{1}{c}{72.10±1.69} \\
          &       & \multicolumn{1}{c}{\textbf{TNMPP}} & \multicolumn{1}{c}{78.61±0.38} & \multicolumn{1}{c}{77.68±0.67} & \multicolumn{1}{c}{77.67±0.22} & \multicolumn{1}{c}{76.73±0.43} \\
          &       & \multicolumn{1}{c}{\textbf{CNNF}} & \multicolumn{1}{c}{76.99±0.31} & \multicolumn{1}{c}{75.98±0.56} & \multicolumn{1}{c}{75.28±1.06} & \multicolumn{1}{c}{73.15±1.88} \\
    \midrule
    \multicolumn{1}{c}{\multirow{15}[6]{*}{\textbf{ARMA}}} & \multicolumn{1}{c}{\multirow{5}[2]{*}{\textbf{Cora}}} & \multicolumn{1}{c}{\textbf{Retrain}} & \multicolumn{1}{c}{74.93±3.94} & \multicolumn{1}{c}{72.63±2.10} & \multicolumn{1}{c}{70.18±1.84} & \multicolumn{1}{c}{61.13±1.63} \\
          &       & \multicolumn{1}{c}{\textbf{Naive}} & \multicolumn{1}{c}{19.03±11.38} & \multicolumn{1}{c}{17.58±9.90} & \multicolumn{1}{c}{12.27±3.61} & \multicolumn{1}{c}{10.73±2.51} \\
          &       & \multicolumn{1}{c}{\textbf{CLR}} & \multicolumn{1}{c}{76.10±0.46} & \multicolumn{1}{c}{75.88±0.47} & \multicolumn{1}{c}{74.08±0.57} & \multicolumn{1}{c}{73.93±0.49} \\
          &       & \multicolumn{1}{c}{\textbf{TNMPP}} & \multicolumn{1}{c}{77.90±0.40} & \multicolumn{1}{c}{77.15±0.64} & \multicolumn{1}{c}{76.28±0.49} & \multicolumn{1}{c}{76.80±0.95} \\
          &       & \multicolumn{1}{c}{\textbf{CNNF}} & \multicolumn{1}{c}{76.57±0.47} & \multicolumn{1}{c}{76.20±0.37} & \multicolumn{1}{c}{74.98±0.74} & \multicolumn{1}{c}{74.55±0.37} \\
\cmidrule{2-7}          & \multicolumn{1}{c}{\multirow{5}[2]{*}{\textbf{Citeseer}}} & \multicolumn{1}{c}{\textbf{Retrain}} & \multicolumn{1}{c}{65.70±0.82} & \multicolumn{1}{c}{63.00±1.45} & \multicolumn{1}{c}{60.73±2.92} & \multicolumn{1}{c}{53.40±1.87} \\
          &       & \multicolumn{1}{c}{\textbf{Naive}} & \multicolumn{1}{c}{19.83±2.83} & \multicolumn{1}{c}{10.33±5.25} & \multicolumn{1}{c}{17.50±0.69} & \multicolumn{1}{c}{14.15±7.72} \\
          &       & \multicolumn{1}{c}{\textbf{CLR}} & \multicolumn{1}{c}{62.53±0.32} & \multicolumn{1}{c}{60.23±0.35} & \multicolumn{1}{c}{59.48±0.65} & \multicolumn{1}{c}{58.10±4.63} \\
          &       & \multicolumn{1}{c}{\textbf{TNMPP}} & \multicolumn{1}{c}{64.50±0.96} & \multicolumn{1}{c}{63.15±0.90} & \multicolumn{1}{c}{62.95±0.93} & \multicolumn{1}{c}{58.95±6.34} \\
          &       & \multicolumn{1}{c}{\textbf{CNNF}} & \multicolumn{1}{c}{63.23±0.21} & \multicolumn{1}{c}{60.70±0.44} & \multicolumn{1}{c}{60.10±0.29} & \multicolumn{1}{c}{58.50±5.13} \\
\cmidrule{2-7}          & \multicolumn{1}{c}{\multirow{5}[2]{*}{\textbf{Pubmed}}} & \multicolumn{1}{c}{\textbf{Retrain}} & \multicolumn{1}{c}{75.03±0.21} & \multicolumn{1}{c}{72.44±0.67} & \multicolumn{1}{c}{68.43±0.54} & \multicolumn{1}{c}{65.97±2.74} \\
          &       & \multicolumn{1}{c}{\textbf{Naive}} & \multicolumn{1}{c}{41.10±0.35} & \multicolumn{1}{c}{27.20±12.60} & \multicolumn{1}{c}{41.15±0.30} & \multicolumn{1}{c}{25.77±13.45} \\
          &       & \multicolumn{1}{c}{\textbf{CLR}} & \multicolumn{1}{c}{74.67±0.67} & \multicolumn{1}{c}{73.76±1.10} & \multicolumn{1}{c}{73.00±1.06} & \multicolumn{1}{c}{71.83±0.64} \\
          &       & \multicolumn{1}{c}{\textbf{TNMPP}} & \multicolumn{1}{c}{76.33±0.15} & \multicolumn{1}{c}{75.70±0.88} & \multicolumn{1}{c}{74.78±0.76} & \multicolumn{1}{c}{73.13±1.18} \\
          &       & \multicolumn{1}{c}{\textbf{CNNF}} & \multicolumn{1}{c}{75.03±0.42} & \multicolumn{1}{c}{74.12±1.26} & \multicolumn{1}{c}{73.75±1.11} & \multicolumn{1}{c}{72.33±0.64} \\
    \midrule
          &       &       &       &       &       &  \\
    \end{tabular}%
    }
  \label{Generalizability}%
\end{table}%

\begin{table*}[tbp]
  \centering
  \caption{Comparative results of our method with GraphEraser \cite{DBLP:conf/ccs/Chen000H022} and GIF \cite{DBLP:conf/www/WuYQS0023} techniques across various dataset partitions.}
    \begin{tabular}{cccccccc}
    \toprule
    \multirow{2}[4]{*}{\textbf{Dataset Division}} & \multirow{2}[4]{*}{\textbf{Model}} & \multicolumn{2}{c}{\textbf{Cora}} & \multicolumn{2}{c}{\textbf{Citeseer}} & \multicolumn{2}{c}{\textbf{Pubmed}} \\
\cmidrule{3-8}          &       & \textbf{Accuracy} & \textbf{Time} & \textbf{Accuracy} & \textbf{Time} & \textbf{Accuracy} & \textbf{Time} \\
    \midrule
    \multicolumn{1}{c}{\multirow{6}[2]{*}{\textbf{\cite{DBLP:conf/www/WuYQS0023,DBLP:conf/ccs/Chen000H022}}}} & \textbf{Retrain} & 88.08±1.00 & 4.09±0.24 & 75.11±0.75 & 4.61±0.08 & 87.52±0.64 & 5.96±0.79 \\
          & \textbf{GraphEraser} & 72.77±1.12 & 1.52±0.13 & 66.76±0.93 & 1.55±0.11 & 76.10±0.36 & 1.60±0.10 \\
          & \textbf{GIF} & 68.71±3.30 & \textbf{0.02±0.00} & 54.17±0.47 & \textbf{0.02±0.00} & -     & - \\
          & \textbf{CLR} & 87.64±0.96 & 0.68±0.16 & 74.29±1.98 & 0.78±0.13 & 87.98±0.65 & \textbf{0.72±0.14} \\
          & \textbf{TNMPP} & \textbf{88.19±1.28} & 0.62±0.02 & \textbf{74.74±1.52} & 0.87±0.23 & 87.96±0.59 & 0.99±0.25 \\
          & \textbf{CNNF} & 87.90±1.02 & 0.72±0.12 & 74.50±1.81 & 0.87±0.16 & \textbf{88.09±0.57} & 0.90±0.17 \\
    \midrule
    \multicolumn{1}{c}{\multirow{6}[2]{*}{\textbf{Our }}} & \textbf{Retrain} & 79.25±0.73 & 4.19±0.43 & 68.06±0.70 & 4.11±0.17 & 75.73±0.71 & 4.46±0.34 \\
          & \textbf{GraphEraser} & 69.34±0.70 & 1.43±0.07 & 65.24±0.68 & 1.48±0.14 & 74.74±0.52 & 1.59±0.07 \\
          & \textbf{GIF} & 59.48±0.94 & \textbf{0.02±0.00} & 34.50±0.29 & \textbf{0.02±0.00} & -     & - \\
          & \textbf{CLR} & 79.63±0.39 & 0.77±0.12 & 68.12±0.29 & 0.66±0.07 & 76.17±0.08 & 0.80±0.15 \\
          & \textbf{TNMPP} & \textbf{80.75±0.47} & 0.91±0.24 & \textbf{69.96±0.59} & 0.71±0.13 & \textbf{78.07±0.50} & 0.82±0.17 \\
          & \textbf{CNNF} & 79.88±0.34 & 0.76±0.10 & 68.44±0.26 & 0.75±0.10 & 76.58±0.21 & \textbf{0.70±0.08} \\
    \bottomrule
    \end{tabular}%
  \label{com_pre}%
\end{table*}%

\subsubsection{Generalizability of Proposed Method}
To assess the generalization capability of our proposed method, we conducted experiments not only on the GCN model but also on the SGC and ARMA models. The experimental results are presented in Tables \ref{MIA results} and \ref{Generalizability}, respectively.

Table \ref{Generalizability} illustrates that the proposed method described in this paper consistently yields models with high model utility across different model architectures. This indicates that our method is capable of accomplishing node unlearning under various architectures with high generality. It's worth noting that the node forgetting achieved by the ARMA model architecture is comparatively lower than the model utility obtained by the GCN and SGC models. However, this discrepancy can be attributed to the inherent architecture of ARMA itself rather than any incompatibility of our method. Additionally, the model utility of the ARMA method under the Retrain approach is also observed to be the lowest.

In Table \ref{MIA results}, we present the unlearning utility under various architectural models. Remarkably, the CNNF method consistently demonstrates the highest unlearning utility across different architectural models. Particularly noteworthy is the observation that CNNF achieves $90\%$ unlearning utility under all architectures when forgetting $80\%$ of the nodes. This highlights the robust generalization capabilities of the method, indicating its efficacy in effectively unlearning nodes across diverse scenarios for privacy protection.

\subsubsection{Comparing with State-of-the-art Method}
In order to assess the state-of-the-art performance of our method, we conducted a comparative analysis with two existing open-source approaches: GraphEraser \cite{DBLP:conf/ccs/Chen000H022} and GIF \cite{DBLP:conf/www/WuYQS0023}. These methods were previously recognized for achieving impressive results in node unlearning. It is worth noting that we omitted methods \cite{DBLP:conf/uss/WangH023} and \cite{10.5555/3524938.3525297}, as the former focuses on inductive GNN models and the latter on linear models, both of which fall outside the scope of our study. Table \ref{com_pre} presents the comparative results of our proposed method with the two existing approaches. The table is divided into two sections based on different dataset divisions. The top section presents the results obtained using the dataset divisions specified in the GraphEraser and GIF papers, while the bottom section presents the results obtained using the dataset divisions employed in our study , with a fixed proportion of $20\%$ for the unlearning data.

As can be seen from the table, the models obtained by our unlearning method outperform the GraphEraser and GIF methods in terms of the model performance and are at least $3\%$ higher, regardless of the dataset and dataset divisions. In the best-case scenario, our method achieves a remarkable $33\%$ improvement. In terms of unlearning efficiency, GIF is the most efficient, but GIF is the lowest in terms of model performance, especially in the configuration of this paper, the model performance of GIF is lower than $60\%$ in all datasets, which surfaces that it destroys the model performance under unlearning. Compared to GraphEraser model, the method in this paper is not only superior in model performance, but also more than $1/3$ faster in efficiency. Overall, combining model performance and efficiency, the method in this paper achieves the best results of graph node unlearning.

Moreover, the table illustrates that the model performance obtained by each method under our dataset divisions is lower than that achieved under the dataset divisions used in GraphEraser and GIF. The difference in performance can be attributed to the dataset division strategies employed in their work. In GraphEraser and GIF, they utilize an 8:2 ratio to re-divide the graph's nodes into the training and testing sets, encompassing all nodes in the graph. This approach ensures a substantial number of labeled nodes in the training set, potentially reaching tens of thousands of nodes, especially in the Pubmed dataset. On the other hand, our dataset divisions involve merging the default training set and the validation set of the dataset to create our training set. As shown in Table \ref{Datasets}, the maximum number of nodes in our training set under this configuration is only $620$, which is significantly smaller than the training set in their dataset divisions. It is widely known in machine learning that larger training datasets tend to result in higher model performance, thereby explaining the observed discrepancy in model performance under the respective dataset divisions.


\subsection{Discussions}

\textbf{Node Number.} Our unlearning work is focused on node-level tasks using a GCN model. Given the default dataset settings, the number of training nodes is relatively small; for instance, the Pubmed dataset contains only $60$ training nodes. Conducting unlearning experiments with $80\%$ of nodes removed under these default settings results in poor and unrealistic performance for the trained models, whether using traditional retraining or our proposed method. To address this limitation and ensure meaningful experimentation, we augmented the training dataset by incorporating a portion of the validation set for our experiments. This approach allows us to evaluate unlearning with a more reasonable number of training nodes, enabling a more accurate assessment of the models' performance after node removal. However, we acknowledge that unlearning experiments with a small number of nodes also hold value and may present different challenges and insights. As such, we plan to explore node unlearning with a few training nodes in our future work.


\textbf{Verified Unlearning.} In this work, we utilized common membership inference attacks to evaluate the effectiveness of unlearning. It's essential to acknowledge that current node-level membership inference attacks may not be flawless and can occasionally result in inference errors. Despite this inherent limitation, we made efforts to bolster the node membership inference attack in our study and achieved accuracy rates of $90\%$ in most cases on the Cora and Citeseer datasets. However, when we applied the same attack to the Pubmed dataset, the accuracy of the membership inference attack dropped below $80\%$ in several instances. This reduced performance of the membership inference attack on Pubmed raises concerns about its efficacy as a validation metric for unlearning. It's noteworthy that membership inference attacks are actively explored in the field of AI privacy research. To ensure a more robust validation of node unlearning, we intend to further investigate and enhance node membership inference attacks in our future work.

\textbf{Finetune vs. SISA.} SISA is a machine unlearning technique that utilizes slicing. However, when the training data is limited, SISA may encounter challenges as each fragment receives insufficient data, resulting in poor performance of the SISA sub-models. Consequently, obtaining satisfactory performance for the final prediction model becomes difficult under such circumstances. Additionally, the SISA technique can be computationally expensive, especially when there is a substantial amount of unlearning data or when the unlearning data is divided into multiple distinct slices. Retraining a large number of sub-models, or possibly all sub-models, can be time-consuming. In contrast, fine-tuning-based methods, like our proposed model, are less affected by the volume of data and demonstrate stable performance during unlearning, even with large amounts of unlearning data. Our method proves to be efficient and effective in maintaining model performance under challenging unlearning scenarios. One notable distinction between the two techniques lies in the level of unlearning achieved. SISA-based methods typically accomplish complete unlearning, removing the influence of unlearning data effectively. On the other hand, assessing the unlearning level in fine-tuning-based methods remains a challenge. Both machine unlearning techniques based on SISA and those based on fine-tuning are still in the early stages of research and warrant further attention and investigation.

\section{Related Work}
\textbf{Machine Unlearning.} Machine unlearning is a technique employed to protect privacy by removing specific training data from a pre-existing model \cite{DBLP:journals/tifs/WangZTYS25, DBLP:journals/nn/ChenZMHL25}. The roots of this research can be traced back to the development of support vector machines (SVM) \cite{DBLP:journals/ml/CortesV95}. Notably, Cauwenberghs et al. \cite{cauwenberghs2000incremental} introduced the Leave-One-Out (LOO) decremental unlearning method, which involves gradually eliminating labeled training data points from an SVM model. Building upon the decremental unlearning approach, Karasuyama et al. \cite{karasuyama2010multiple} expanded the methodology by incorporating multi-parameter programming. This enhancement enables the simultaneous addition and/or removal of multiple data points from the model. In further advancements, Cao et al. \cite{DBLP:conf/sp/CaoY15} explored machine unlearning using a summation-based methodology. In this approach, model updates and sample removal are achieved by subtracting the relevant summations.


In recent years, deep neural network unlearning has been a topic of active investigation among researchers. In the field of machine unlearning, Bourtoule et al. \cite{DBLP:conf/sp/BourtouleCCJTZL21} introduced the Sharded, Isolated, Sliced, and Aggregated (SISA) framework. This framework leverages the principles of Sharded, Isolated, Sliced, and Aggregated training to achieve data unlearning. Notably, it offers the advantage of accelerating the training process by intentionally reducing the influence of specific data points. In the context of recommender systems, Chen et al. \cite{DBLP:conf/www/Chen0ZD22} applied the SISA framework and presented a comprehensive and effective machine learning approach for recommendation tasks. Their work highlights the broad applicability of SISA in various domains, including recommender systems. Additionally, Aldaghri et al. \cite{DBLP:journals/access/AldaghriMB21} explored the application of SISA-based coded machine unlearning. Their research focuses on leveraging coding techniques to enhance the efficiency and security of machine unlearning processes. Collectively, these studies contribute to the advancement of machine unlearning techniques and showcase the potential of the SISA framework in different application scenarios. 

Many researchers have been actively investigating machine unlearning approaches that involve fine-tuning models. In the context of machine unlearning, Graves et al. \cite{DBLP:conf/aaai/GravesNG21} have proposed two efficient techniques for model fine-tuning. Their solutions aim to preserve the model's validity while protecting it against model inversion and membership inference attacks. By allowing model owners or data holders to remove private information from the model, their approach addresses privacy concerns effectively. Wu et al. \cite{DBLP:conf/icml/WuDD20} have introduced a machine unlearning technique that enables fast retraining of machine learning models using cached data from the training phase. This approach focuses on optimizing the retraining process by leveraging the existing cached data, thereby reducing the computational burden associated with retraining from scratch. In a different vein, Wu et al. \cite{DBLP:conf/aaai/WuHS22} have proposed a novel framework called performance unchanged model augmentation. This framework explicitly models the impact of each training data point on the model's generalization capacity, considering multiple performance criteria. By optimally re-weighting the remaining data after removing labeled data, the model is fine-tuned to mitigate the negative effects of data removal. These research contributions provide valuable insights into machine unlearning and offer practical techniques for fine-tuning models to enhance privacy, computational efficiency, and model performance. The continuous progress in this field indicates the growing importance of developing robust and effective machine unlearning methods to address the challenges posed by data privacy and model security in machine learning applications.

\textbf{Machine Unlearning on GNNs.} 
Graph Neural Networks (GNNs) \cite{DBLP:journals/tnn/ScarselliGTHM09} have gained significant popularity as machine learning models designed specifically for handling graph-structured data. However, like other models, GNNs are susceptible to privacy attacks and potential leakage of sensitive graph data information during deployment \cite{DBLP:journals/nn/LiLGLQW25}. Consequently, researchers have focused on studying methods for GNN unlearning to enhance the privacy and security of graph data. 

Similarly, researchers have explored the application of the SISA method for unlearning in GNNs. The direct application of SISA to graphs could potentially compromise the structural information of the graph and lead to reduced model utility. Chen et al. \cite{DBLP:conf/ccs/Chen000H022} proposed an SISA unlearning framework specifically designed for graph data. Their approach introduced two novel graph partitioning algorithms and a learning-based aggregation method, addressing the challenge of directly applying SISA to graph data. Nevertheless, it's important to note that this method does lead to a certain extent of performance degradation in the model. In contrast, the performance of all three methods proposed in this paper after executing node unlearning surpasses the performance of the method presented by Chen et al. Moreover, our method also exhibits greater efficiency in terms of unlearning speed compared to their method.

There have been several studies on graph node unlearning based on fine-tuning. Chien et al. \cite{DBLP:journals/corr/abs-2206-09140} introduced the first known framework for certified graph unlearning of GNNs. They considered three different types of unlearning requests, including node features, edges, and node unlearning. However, it is important to note that their framework was specifically tailored for linear models, which may limit its applicability to more complex GNNs. On the other hand, Wu et al. \cite{DBLP:conf/www/WuYQS0023} explored node unlearning by considering neighboring nodes in the graph. They investigated the influence function by incorporating a loss term that accounts for the structural dependence of the affected neighborhood into the objective of the traditional influence function. This additional loss term provides a deeper understanding of the unlearning mechanism and enables the derivation of closed-form solutions for parameter variations. Their method demonstrates efficiency in the context of node unlearning. However, it is crucial to acknowledge that unlearning has a significant impact on the model's performance, as evident from the results presented in Table \ref{com_pre}, when their approach is applied. In comparison, the method proposed in this paper sustains the superior performance of the model even after the node unlearning process.

\section{Conclusion}
In this paper, we propose three novel node unlearning methods aimed at efficiently removing training nodes from GNNs. The first method, Class-based Label Replacement, draws inspiration from machine unlearning techniques commonly employed in image and related domains. This method performs unlearning by replacing the target class of the unlearning nodes with the mean posterior probability of the same class of nodes in the testing set. The remaining two methods, Topology-guided Neighbor Mean Posterior Probability and Class-consistent Neighbor Node Filtering, leverage the topological features of graph data to achieve effective node unlearning. Additionally, we introduce a node membership inference attack method to evaluate the utility of node unlearning. Through conducting experimentation on three diverse datasets, our results demonstrate that all three proposed methods effectively achieve the efficient removal of training nodes in GNNs.

\bibliographystyle{cas-model2-names}

\bibliography{ref}



\end{document}